%% file: emnlp2023.tex
\pdfoutput=1

\documentclass[11pt]{article}

\usepackage[]{EMNLP2023}

\usepackage{times}
\usepackage{latexsym}

\usepackage[T1]{fontenc}

\usepackage[utf8]{inputenc}

\usepackage{microtype}

\usepackage{inconsolata}

\newcommand{\sllm}{\mathbf{SLLM}}
\newcommand{\bs}{\backslash}
\newcommand{\semantics}[1]{[\![ #1 ]\!]}

\usepackage{tikz}
\input{tikzpreamble.tex}

\title{Towards Transparency in Coreference Resolution: \\ A Quantum-Inspired Approach}

\author{Hadi Wazni \\
  University College London \\
  \texttt{hadi.wazni.20@ucl.ac.uk} \\\And
  Mehrnoosh Sadrzadeh \\
  University College London  \\
  \texttt{m.sadrzadeh@ucl.ac.uk} \\}

\begin{document}
\maketitle
\begin{abstract}

Guided by grammatical structure, words compose to form sentences, and guided by discourse structure, sentences compose to form dialogues and documents. The compositional aspect of sentence and discourse units is often overlooked by machine learning algorithms. A recent initiative called Quantum Natural Language Processing (QNLP) learns word meanings as points in a Hilbert space and acts on them via a translation of grammatical structure into Parametrised Quantum Circuits (PQCs). Previous work extended the QNLP translation to  discourse structure using points in a closure of Hilbert spaces. In this paper, we evaluate this translation on a Winograd-style pronoun resolution task. We train a Variational Quantum Classifier (VQC) for binary classification and implement an end-to-end pronoun resolution system. The simulations executed on IBMQ software converged with an F1 score of 87.20\%. The model outperformed two out of three classical coreference resolution systems and neared state-of-the-art SpanBERT. A mixed quantum-classical model yet improved these results with an F1 score increase of around 6\%.

\end{abstract}

\section{Introduction}

Large language models (LLMs), such as GPT-3 \cite{brown2020language}, have achieved impressive success in various NLP tasks and have become increasingly common in everyday life through search engines, personal assistants, and other applications. They are trained on vast corpora of text, which are sourced from books, articles, and websites. LLMs learn complex connections between words and phrases by predicting the likelihood of a word appearing in the context of other words. These learned probability distributions capture the statistical patterns of word co-occurrences in data; due to this, LLMs are also known as distributional language models.

Despite their successes in advancing language understanding and generation, LLMs often face criticism for being black boxes \cite{buhrmester2019analysis}. This means that it is challenging to understand how they make their predictions, which can in turn make them unreliable and difficult to debug. One way to enhance the  transparency and interpretability of these models  is to explicitly integrate linguistic structure \cite{doi:10.1080/00029890.1958.11989160, Chomsky57a} into them. 

A notable approach attempting this integration is the Distributional Compositional Categorical  (DisCoCat) model \cite{Coeckeetal2010, kartsaklis-sadrzadeh-2013-prior}, which pioneered the paradigm of merging explicit grammatical (or syntactic) structure with distributional (or statistical) data for encoding and computing meanings of sentences. DisCoCat offered tools for a compositional statistical modelling of sentence-level linguistic phenomena, such as lexical entailment and ambiguity, by providing transparent meaning assignments for complex syntactic structures, e.g.  relative and possessive clauses \cite{Sadrzadeh_2013, Sadrzadeh_2014}, conjunctive and negation operations \cite{lewis2020logical}. Its underlying theory, however, relied on generalisations of vectors to higher order tensors, which made the framework in need of large  computational resources and led to limited scalability.

Conversely, tensors are natural components of quantum systems, and quantum computing resources can efficiently learn them. This idea has led to the development of Quantum Natural Language Processing (QNLP). In QNLP, words are represented as points within a Hilbert space, grammatical structures are represented as Parameterised Quantum Circuits (PQCs), and the learning of circuit parameters is achieved through simulations conducted on accessible quantum computing resources, such as IBMQ quantum computers.
QNLP has so far been applied to a variety of tasks, e.g.  sentence classification \cite{QnlpInPractice}, sentence generation \cite{karamlou2022quantum}, question answering \cite{Meichanetzidis_2023}, sentiment analysis \cite{9951286, stein2023applying, ganguly2023quantum}, musical composition \cite{miranda2021quantum}, and language translation \cite{abbaszade2023quantum}. Moreover, the theoretical underpinnings of QNLP have been extended to model discourse structure and have been tested on a limited toy dataset \cite{wazni2022quantum}.
 
In this paper, we expand this dataset by introducing a few-shot prompting technique and generate synthetic Winograd-style ambiguous coreference sentences \cite{levesque2012} using GPT-3. We apply this method to a set of initial sentences from \cite{rahman-ng-2012-resolving} and create a dataset consisting of 16,400 entries. This dataset have a larger number of data points, longer and more complex sentences, and a broader range of grammatical structures when compared to the dataset in \cite{wazni2022quantum}, where sentences followed a subject-verb-object structure.

We train a Variational Quantum Classifier (VQC) for binary classification and integrate it into an end-to-end pronoun resolution system. Our system's performance surpasses that of classical coreference resolution systems such as CoreNLP \cite{manning-etal-2014-stanford} and Neural Coreference \cite{clark-manning-2016-deep,clark-manning-2016-improving}, and it achieves results that are close with the state-of-the-art SpanBERT \cite{lee2018}, with an F1 score of 87.20\%. 
Following recent practice in quantum machine learning (QML) \cite{qensemble1,qensemble2}, we merge our quantum system with classical engines to construct a \emph{mixed quantum-classical} pronoun resolver. In alignment with results observed in QML across various domains \cite{qfraud,qdrug,qMNIST}, we find that the classical and quantum results are \emph{complementary}, thus our mixed approach yields a significant performance improvement, resulting in an approximate 6\% increase in the F1 score.

\section{Background and Related Work}

In the DisCoCat framework, the grammatical structure of a sentence guides the composition of its word-meanings, leading to the derivation of meaning for the sentence as a whole \cite{coecke2020foundations,Coeckeetal2013}. The  grammatical structures are modelled by proofs derived using the rules of  Joachim Lambek’s logic of syntax, known as the Lambek Calculus \cite{Lambek1988}. These  proofs are interpreted as \emph{processes} and modelled by morphisms of a  monoidal category, which comes equipped with a \emph{string diagrammatic} graphical notation \cite{piedeleu2023introduction}. Examples of processes that can be effectively modelled by a monoidal category include linear maps over finite-dimensional vector spaces, and this was the initial concept behind the introduction of DisCoCat. Atomic words like noun phrases are represented as points within finite-dimensional vector spaces, while functional words such as adjectives and verbs are depicted as points within the tensor products of these vector spaces. The interconnection of vector and tensor spaces is facilitated through their grammatical dependencies. By contracting these dependencies, the framework allows for the derivation of the overall meaning of the entire sentence.

In fact, the formulation of vectors and tensors into a monoidal category goes back to a framework known as categorical quantum mechanics (CQM), which reformulated quantum theory in terms of process theories and used string diagrams to describe quantum protocols  \cite{abramsky2008categorical, coecke_kissinger_2017}. For a  detailed introduction to quantum computing and CQM, see \cite{nielsen_chuang_2010, coecke_kissinger_2017, sutor2019dancing}. As a result, monoidal categories and string diagrams became a common base in which one can use analogical reasoning to relate language with quantum theory. For instance, Hilbert spaces, where quantum states are encoded, are vector spaces, so quantum states are related to word-meanings and grammatical reductions correspond to processes such as quantum maps, quantum effects, and measurements. 

\subsection{Lambek Calculus and its modal extensions}

\begin{figure*}
\centering
\resizebox{1\textwidth}{!}{%
	{%
\beginpgfgraphicnamed{translation}
\InputIfFileExists{translation.tikz}{}{\input{./tikz/translation.tikz}}
\endpgfgraphicnamed}
}%
\caption{Translation from string diagrams to PQCs using a single-layer IQP ansatz, where each grammatical type is mapped to a 1-qubit space.}
\label{fig:transform}
\end{figure*}
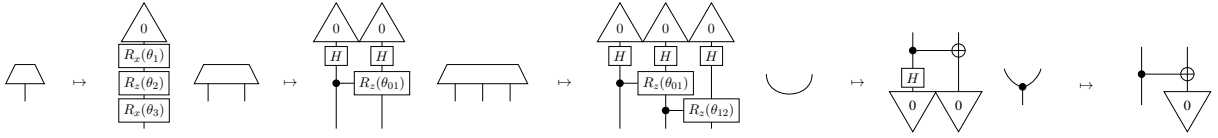

The formulae of \emph{Lambek Calculus} (LC) are generated according to the following BNF:

\[A,B::= A\in At \mid A\cdot B \mid A\backslash B \mid A/B \]

\noindent
Atomic types $A\in At$ are atomic linguistic types, e.g. noun phrases $n$ and sentences $s$,  multiplication $A\cdot B $ is their composition, and the slashes $A \backslash B$ and $A/B$ build complex types, e.g. for words with function types such as adjectives and verbs.  

In \cite{Kanovich2020}, an extension of LC with two operations $!A$ and $\nabla A$ was introduced.  The new logic was  named  \emph{Lambek calculus with soft sub-exponentials} ($\mathbf{SLLM}$).   In \cite{mcpheat2021LACL},  the new modal formulae were used to model the linguistic types found in discourse, e.g. pronouns and other ellipsis markers.  The $!$-modal types were used for copying referents up to a bound $k$, and  the $\nabla$-modal types  moved them  to the locations of their markers, where they were referred to.   The authors showed how the logic could model and reason about definite pronoun discourse ambiguities, such as the Winograd schema examples, and  sloppy vs strict readings of elliptic sentences. 

In \cite{Coeckeetal2013},  the following vector space semantics was proposed for LC:
 \vspace{-2mm}  
\[
\semantics{A} = V \in FdVect, \hspace{5mm} \semantics{A \cdot B} = \semantics{A} \otimes \semantics{B}
\]
\[
 \semantics{A \bs B} = \semantics {B/A} = \semantics{A}^* \otimes \semantics{B}
\]

\noindent 
In this semantics, atomic linguistic types are interpreted as finite-dimensional vector spaces and their multiplication as the tensor product of spaces; the  slash types are interpreted as  the set of all linear maps between their two spaces, via the dual vector space denoted by $(-)^*$. Words are interpreted as elements of the vector spaces associated to their types. This semantics was extended to $\mathbf{SLLM}$ in \cite{mcpheat2021LACL}, by interpreting the copiable linguistic categories  as $k$-\emph{truncated} Fock spaces, defined as follows:
 \vspace{-2mm}
\[
\semantics{!^k A} = \bigoplus_{i=0}^{k}\semantics{A}^{\otimes i} = k \oplus \semantics{A} \oplus (\semantics{A} \otimes \semantics{A})  \cdots
 \]
 \vspace{-2mm}
\[
 \cdots \oplus (\semantics{A}\otimes \semantics{A}\otimes \semantics{A}) \oplus \cdots \oplus\semantics{A}^{\otimes k}
 \]

A Fock space closes its base vector $A$ under an infinite number of tensor products, and a $k$-truncated version of it only looks at the first $k$ tensors. Access to any copies of a linguistic category (less than the bound $k$) is facilitated by projecting to that  layer. Movable categories take advantage of the commutativity of the tensor product between finite-dimensional vector spaces. The direct sum operation $\oplus$ cannot be directly represented using the quantum gates available in QNLP, which corresponds to the gates provided by IBMQ. We thus translate it into a PQC after projecting it to the desired layer. 

A summary of the translation between our Fock space semantics and PQC is provided in Figure \ref{fig:transform}. Due to space restrictions, we present the translation for the case where only a single qubit is allocated  to each atomic linguistic type. In theory, the translation is easily extendible to larger numbers of qubits, but in practice one will face computational limitations. 
There are two types of diagrams: those on the left, which represent string diagrams associated with vector spaces, and the ones on the right, which depict diagrams used for quantum circuits. On the string diagrammatic side, a parallelogram box with one leg depicts words with an un-copied atomic types. A parallelogram with many legs either depicts a words with a copied type or a functional type. Cupped lines depict the application of a linear map. The concatenation of two atomic sentence types has a conjunctive (rather than tensorial) interpretation, and this is modelled by the Frobenius multiplication between vector spaces. This multiplication is diagrammatically denoted by a bullet symbol ($\bullet$). 

In Figure \ref{fig:ex1_a}, an example of a string diagram, where \textit{``books''} and \textit{``learning''} are depicted without being copied, which is indicated by their parallelograms having one leg each. \textit{``The students''} is copied and has a parallelogram with two legs. The pronoun \textit{``They''} is shown with one input and one output, giving it two legs. The verbs  \textit{``were''} and \textit{``read''} are represented with two inputs and one output, resulting in three legs each. Cupped lines in the diagram illustrate the application of verbs to their subjects and objects, while a bullet symbol ($\bullet$) is used to connect \textit{``The students read the books''} with \textit{``They were learning''}.

On the circuit side, a triangle labeled with $0$ represents a qubit state in the zero computational basis. A box labeled with $H$ signifies a Hadamard gate. A CNOT gate is denoted by a dot connected horizontally to $\oplus$. A controlled-Z-rotation gate with angle $\alpha$, depicted as a box labeled with $R_{\alpha}(\theta_i)$, is connected horizontally to a control qubit, where $\alpha$ can be $x$, $y$, or $z$, and $\theta$ ranges from $0$ to $2\pi$. An upside-down triangle labeled with $0$ signifies a measurement in the computational basis, post-selected to be zero. 

\section{Methodology}

We build upon the steps in \cite{QnlpInPractice} to represent an entire discourse as a PQC.

\paragraph{Parsing and Diagram Generation:} The first step involves parsing a discourse into a proof in $\mathbf{SLLM}$. We do this via a  translation to Combinatory Categorial Grammar (CCG)\footnote{A grammar formalism inspired by combinatory logic and developed in \cite{10.7551}}, which enables the use of the state-of-the-art parser \cite{clark2021old, yeung2021ccgbased}. The parse trees are then transformed to string diagrams through \textit{DisCoPy} \cite{de_Felice_2021}.

\paragraph{Diagram Optimisation:}

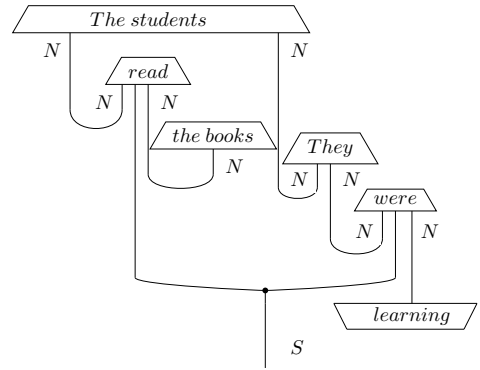
\begin{figure}[h]
\hspace{0.7cm}
\resizebox{0.4\textwidth}{!}{%
      {%
\beginpgfgraphicnamed{example1_c}
\InputIfFileExists{example1_c.tikz}{}{\input{./tikz/example1_c.tikz}}
\endpgfgraphicnamed}
    }%
\caption{An optimized $\mathbf{SLLM}$ diagram for a pair of sentences \textit{``The students read the books. They were learning.''} To enhance clarity, we treat the determiner-noun phrases \textit{``The students''} and \textit{``The books''} as single units, as determiners are eventually discarded in the rewriting process.}
\label{fig:ex1_a}
\end{figure}

The number of qubits available on contemporary quantum computers is  restricted. For instance, IBM's largest superconducting quantum computer, as of now, has a maximum of 433 qubits\footnote{https://www.ibm.com/quantum/roadmap}. Publicly accessible devices typically offer  fewer qubits, often less than 10. Consequently,  in the second step, the string diagrams are optimised to  minimise the number of qubits associated to them after the  translation.   QNLP diagrams are composed of a layer of tensors, followed by a layer of applications between the tensors. 
 %
%
One approach to reduce the number of qubits is  elimination of cups through the transformation of states into effects. Another approaches aims for stretching and reordering them. \textit{Lambeq} \cite{lambeq_paper} supports additional rewriting rules. An example of an optimised diagram is provided in Figure \ref{fig:ex1_a}. 


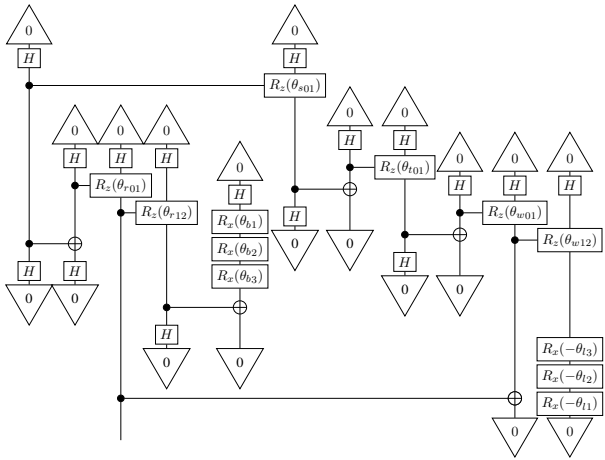
\begin{figure}[h]
\resizebox{0.5\textwidth}{!}{%
      {%
\beginpgfgraphicnamed{example1_d}
\InputIfFileExists{example1_d.tikz}{}{\input{./tikz/example1_d.tikz}}
\endpgfgraphicnamed}
    }%
\caption{A PQC using the IQP ansatz, transformed from the string diagram presented in Figure \ref{fig:ex1_a}. The parameters $\{\theta_{s01}\}$, $\{\theta_{r01}, \theta_{r12}\}$, and $\{\theta_{b1}, \theta_{b2}, \theta_{b3}\}$ are associated with the terms \textit{the students}, \textit{read}, and \textit{the books} respectively, while $\{\theta_{t01}\}$, $\{\theta_{w01}, \theta_{w12}\}$ and $\{-\theta_{l3}, -\theta_{l2}, -\theta_{l1}\}$ are associated with \textit{They}, \textit{were}, and \textit{learning} respectively.}
     \label{fig:ex1_b}
\end{figure}

\paragraph{Quantum Circuit Transformation:} In the last step, the optimised string diagrams are transformed into quantum circuits. This conversion relies on a parameterisation scheme, known as an \textit{ansatz}. An \textit{ansatz} serves as a mapping that determines the quantity of qubits linked with each wire in the string diagram, along with a distinct variational quantum circuit associated with each word. In this study, we choose the popular \emph{Instantaneous Quantum Polynomial} (IQP) ansatz, developped in \cite{Shepherd_2009, Havl_ek_2019}. The resulting  quantum circuits are ready for execution on either a quantum computer or a simulator. The details of  training these circuits can be found in Section \ref{training_section}.
Figure \ref{fig:ex1_b} illustrates the circuit derived from the diagram presented in Figure \ref{fig:ex1_a}.

\begin{figure*}
  \centering
  \begin{minipage}[b]{0.45\textwidth}
    \centering
    \resizebox{\textwidth}{!}{%
      {%
\beginpgfgraphicnamed{example2_c}
\InputIfFileExists{example2_c.tikz}{}{\input{./tikz/example2_c.tikz}}
\endpgfgraphicnamed}
    }%
  \end{minipage}%
  \begin{minipage}[b]{0.45\textwidth}
    \centering
    \resizebox{\textwidth}{!}{%
      {%
\beginpgfgraphicnamed{example2_d}
\InputIfFileExists{example2_d.tikz}{}{\input{./tikz/example2_d.tikz}}
\endpgfgraphicnamed}
    }%
    \label{fig:ex1_d}
  \end{minipage}
\caption{An optimised $\mathbf{SLLM}$ diagram where the pronoun refers to the object: \textit{``The students read the books. They were interesting.''} The diagram along with its transformation into a PQC.}
\end{figure*}
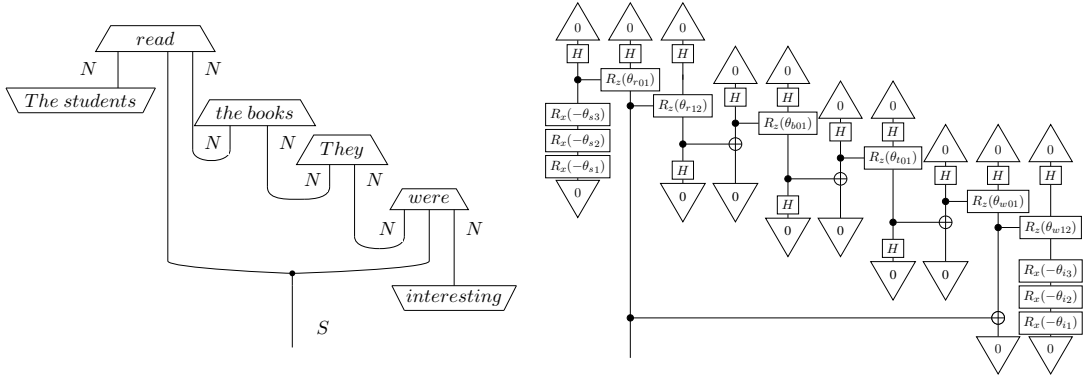

\section{Classification Task}

Pronoun resolution is a computational linguistic process that involves identifying the antecedent of a pronoun within a text. 
In our experiment, we consider pronoun resolution as a supervised binary classification task. Given a sentence containing a pronoun, the goal is to determine whether a potential antecedent (such as a noun or noun phrase) in the preceding sentence is the correct referent for the pronoun or not. This task requires training a variational quantum classifier with labeled data, where each pronoun-noun pair is classified as  \textit{non-coreferent} or \textit{coreferent}. The code and data used in this paper are available at the following link: \url{https://github.com/hwazni/Qcoref}


\subsection{Dataset}

The process of training PQCs involves optimising multiple parameters associated with each word in a given dataset, with the objective of minimising the loss value on the training set. When it comes to predicting the output for a test sample, a PQC is constructed based on the input sentence. Each word in the sentence is associated with a specific set of parameters learned during the training process. A significant challenge arises when an out-of-vocabulary word is encountered during inference, which includes testing or using the model for predictions. These words lack a predefined parameter assignment. To address this issue, there are several approaches, including random initialisation, replacement with a special token like ``UNK'' for unknown words, or establishing an overlap between the test and training vocabularies. 
In our case, we fix  a set of words with grammatical relations between them, then use these and prompt the GPT-3 model to generate pairs of sentences that exhibit a substantial overlap in vocabulary.


\begin{table*}
\centering
\begin{tabular}{lccccc}
\hline
\textbf{Sentence Pair $(S_1, S_2)$}& \textbf{Pronoun} & \textbf{Noun} & \textbf{Label}\\
\hline
The students researched the books. \\They were seeking new insights. & They & students & 1\\
\hline
The massive storm cancelled the flight. \\ It was full of passengers. & It & storm & 0 \\
\hline
The precise sniper eliminated the ruthless terrorist. \\ He was a vicious dealer. & He & terrorist & 1 \\
\hline
The exhausted sailors threw themselves off the boats. \\ They were in poor condition. & They & sailors & 0 \\
\hline
\end{tabular}
\caption{\label{dataset-table}
Dataset entries: each sentence pair is labeled with a ``0'' signifying that the pronoun do not refer to the candidate noun. Conversely, a ``1'' label indicates that the pronoun and the noun are co-referential.
 }
\end{table*}

In the initial step, we selected entries from the definite pronoun resolution dataset introduced in \cite{rahman-ng-2012-resolving}, an extension of the Winograd Schema Challenge dataset \cite{levesque2012}. We excluded sentences containing proper nouns and negation, and gave preference to shorter sentences. This process resulted in a total of 10 entries. Each entry was a pair of sentences. The first sentence, exemplified by $E_1:$ \textit{The students read the books}, contains two referent nouns, namely, \textit{the students} and \textit{the books}. In the second sentence, an ambiguous pronoun is introduced, referring to one of the referents in $E_1$. For instance, it could be either $E_2:$ \textit{They were learning} or \textit{They were interesting}. Notably, the pronoun aligns with gender, number, and semantic class concerning each of the candidate referents mentioned in the first sentence. For each initially selected pair $(E_1, E_2)$, we created an additional set of pairs $(S_1, S_2)$ incorporating a more diverse range of grammatical structures. In these template pairs, $S_1$ retained the same referents as $E_1$, and $S_2$ maintained the same co-reference relation with $E_1$. Below is the list of template pairs for the \textit{student-book} example.




\noindent
\begin{enumerate}
\item The students \textit{(verb, phrasal verb, verb phrase)} the books. They were \textit{(adjective, gerund phrase)}.
\item The \textit{(adjective)} students \textit{(verb, phrasal verb, verb phrase)} the books. They were \textit{(adjective, gerund phrase)}.
\item The students \textit{(verb, phrasal verb, verb phrase)} the \textit{(adjective)} books. They were \textit{(adjective, gerund phrase)}.
\item The \textit{(adjective)} students \textit{(verb, phrasal verb, verb phrase)} the \textit{(adjective)} books. They were \textit{(adjective, gerund phrase)}.
\end{enumerate}

\noindent
The templates replace the verb \textit{``read''} by another \textit{verb}, \textit{phrasal verb} or a \textit{verb phrase}. Similarly, the adjectives \textit{``learning''} and \textit{``interesting''} can be replaced by another \textit{adjective} or \textit{gerund phrase}. Sample templates for different examples are listed in section \ref{sec:appendix}.

Next, we utilize the prompt provided in the box below in GPT-3 along with template pairs. This technique referred as few-shot prompting, where we provide examples in the prompt to steer the model to better performance \cite{brown2020language, kaplan2020scaling, touvron2023llama}. 
Note that the red tokens are modified for each example.\\

\noindent
\fbox{\parbox{\linewidth}{\textbf{Provide alternative sentences by replacing the words or phrases inside the brackets for each statement. Utilize different {\color{red}verbs}, {\color{red}phrasal verbs}, {\color{red}verb phrases}, {\color{red}adjectives}, or {\color{red}gerund phrases} to create new sentences based on the given structure. Ensure that the pronoun  {\color{red}`they'} in the second sentence refers to {\color{red} `students'} / Ensure that the pronoun  {\color{red}`they'} in the second sentence refers to  {\color{red}`books'}}}}\\

From the GPT-generated output, we eliminated incorrect referent sentences and duplicate examples, retaining only well-formed sentences that possess meaningful content. We carefully handpicked between 300 to 400 examples for each entry, ensuring a balanced distribution of pronoun references. Then we used the generated linguistic elements, including \textit{verbs, phrasal verbs, adjectives, adverbs, nouns, compound nouns, verb phrases, adverbial phrases, gerund phrases, and prepositional phrases}, with 8 distinct structural patterns to generate over 8 million diverse combinations. We randomly choose 1800 pairs for each example, with one example with 200 pairs. This ended up with 16,400 (0.2\%) examples, comprising approximately 200,000 words, with 1,214 unique vocabulary. Through this approach, we achieved the generation of coherent sentences that uphold grammatical correctness and preserve semantic consistency, as a result a high quality was ensured for the dataset. The dataset was subsequently split into three subsets: 10,496 pairs ($\sim$60\%) for training, 2,624 pairs ($\sim$20\%) for validation, and 3,280 pairs ($\sim$20\%) for testing. The training and testing datasets share a common vocabulary of 95\%, while none of the sentence pairs in the testing set appears in the training or validation sets. Some examples of the dataset are provided in Table \ref{dataset-table}.

\subsection{Simulating the quantum circuits}

Computation using currently available quantum computers, which are called NISQ for Noisy Intermediate-Scale Quantum,  is slow, noisy and limited. They lack the practicality needed for extensive training and comprehensive comparative analyses \cite{Preskill_2018}. For this reason, and especially at the early stages of modelling, proofs-of-concept are obtained by running simulations. 
A simple way to simulate a quantum computation is to use linear algebra; since quantum gates correspond to complex-valued tensors, each circuit can be represented as a tensor network where computation takes place as a result of a series of tensor contractions. The output of these contractions  is the ideal probability distribution of the measurement outcomes on a noise-free quantum computer, i.e.  an idealistic approximation of the sampled probability distribution obtained from a NISQ device.  We conduct our experiments using noiseless non-shot-based simulations  utilizing the \textit{NumPyModel} of \textit{Lambeq} \cite{kartsaklis2021lambeq}  with a JAX backend \cite{47008}.

\subsection{Training}
\label{training_section}

We implement a  hybrid classical-quantum training approach in which the quantum computer is responsible for computing the meaning of the sentence by connecting the quantum states in a quantum circuit and the classical computer is used to calculate the training's loss function. During each iteration, a new set of quantum states is generated, driven by the loss function's outcome from the preceding iteration. This iterative procedure ensures that the quantum states are continually refined to enhance the model's performance and accuracy.

\input{graph_results.tex}

%
%

Specifically, the sentence pair $(S_1, S_2)$ within each dataset entry are combined to create a single output quantum state. These resultant states are the inputs to our binary classifier. In principle, they can be any quantum map that take two sentences as input and produce a sentence as the output (recall the whole circuit is represented by an open sentence wire). A CNOT gate is used to combine the two sentences, as it encodes a commutative Frobenius multiplication ($\bullet$) and acts similar to a logical conjunction. The resulting quantum circuit is denoted by $S_1 \bullet S_2$ and evaluated for an initial set of parameters $\Theta = \left ( \theta_1, \theta_2, ..., \theta_k \right )$ on a quantum computer giving an output state $\left | S_1 \bullet S_2 \left ( \Theta \right ) \right\rangle$. The expected prediction is given by the Born rule, i.e. as follows:

\[
l^i_\Theta (S_1 \bullet S_2) := \left | \langle i | S_1 \bullet S_2 (\Theta) \rangle \right |^2 + \epsilon
\]

\noindent
where, $i \in \left \{ 0, 1 \right \}$,  $\epsilon$ is a smoothing term with the value $10^{-9}$, 
and $l_\Theta (S_1 \bullet S_2)$ is the following probability distribution:
\[
l_\Theta (S_1 \bullet S_2) := \frac{(l^0_\Theta (S_1 \bullet S_2), l^1_\Theta (S_1 \bullet S_2))}{\sum_{i}^{} l^i_\Theta (S_1 \bullet S_2)}
\]

The predicted label is obtained by rounding the probability distribution  to the nearest integer $\lfloor l_\Theta (S_1 \bullet S_2) \rceil$ and represented as one-hot encoding. This means if $\lfloor l_\Theta (S_1 \bullet S_2) \rceil \l < 0.5$,  the predicted label $[0, 1]$ corresponds to \textit{non-coreferent} mentions, and if $\lfloor l_\Theta (S_1 \bullet S_2) \rceil \geq 0.5$, the predicted label  $[1, 0]$ corresponds to \textit{coreferent} mentions.

To find the optimal parameters for our model, the predicted label is compared with the training label using a binary cross-entropy loss function and minimised using a non-gradient-based optimisation algorithm known as SPSA (Simultaneous Perturbation Stochastic Approximation) \cite{705889}. 

For the hyper-parameters, we set the initial learning rate \textit{a} to 0.1, the initial parameter-shift scaling \textit{c} to 0.06, and the stability constant \textit{A} to 20. We run for 2000 epochs of SPSA during which we evaluate the training loss and accuracy. This process is repeated 15 times with random seed values. This is essential since the gradient computed by the SPSA procedure is an approximation and the performance in QML is known to be very sensitive to the initial parameter assignment \cite{PRXQuantum, Grant_2019, McClean_2018}.

\section{Results and Discussion}
\subsection{Quantum Approaches: SLLM vs Bag-of-Words}
The graphs in Figure \ref{training_per} illustrate how the models converged smoothly. Across different runs, a common trend emerges—training loss decreases and training accuracy increases steadily. Initially, the average training loss is 1.144, which drops to 0.483 after 2000 iterations. Minimum and maximum values range from 0.369 to 0.571 for different runs. Similarly, the average training accuracy starts at 0.514 and ends at 0.752 after 2000 iterations. The highest recorded accuracy is 0.827, and the lowest is 0.682 amongst all  the runs. 
The testing accuracy rates vary between 0.628 and 0.782, averaging around 0.70. 
These results demonstrate that the model is able to generalise its predictions beyond training, with well-balanced performance levels. 

To understand whether the promising performance of the $\mathbf{SLLM}$ classifier is due to the structural symbolic type-driven representations or the use of PQCs, we conducted a comparative analysis with quantum circuits generated from a simple bag-of-words diagram (see section \ref{sec:appendix}). In this approach, each word is represented with a single qubit, regardless of its grammatical type (e.g., noun, adjective, or verb). Consequently, this model disregards sentence structure and connects all qubits using CNOT gates (the simplest counterparts to addition in quantum circuits). We trained the model under identical hyper-parameters and the same number of training runs. However, its performance fell short, yielding an average testing accuracy of 0.557.

\subsection{Classical Approaches: SVM, CoreNLP, Neural Coreference, SpanBERT}

We implemented a Support Vector Machine (SVM) for a binary classification task and evaluated its performance in comparison to our VQC. The inputs to the SVM were pre-trained Sentence-BERT embeddings \cite{reimers2019sentencebert}, one per each dataset entry. We also experimented with a compositional model, by adding SBERT word embeddings of each entry, as shown below: \\



\noindent
\textbf{SVM Full} : $\overrightarrow{E}$ \\
\textbf{SVM Add} : $(\overrightarrow{w_1}+\overrightarrow{w_2}+\overrightarrow{w_3}...) + (\overrightarrow{w_1}+\overrightarrow{w_4}+\overrightarrow{w_5}...)$\\

%
%

\noindent
In the above, $E$ is an entry such as: \textit{``The students researched the books. The students were seeking new insights."} labeled as $1$ or \textit{``The massive storm cancelled the flight. The storm was full of passengers."} labeled as $0$. 
In \textbf{SVM Add}, \overrightarrow{w_1} is a candidate referent, e.g.  \textit{students} or \textit{storm},  and \overrightarrow{w_2}, \overrightarrow{w_3}, \overrightarrow{w_4}, \overrightarrow{w_5} are all the other words. 

The objective here was to assess the discourse relation within each entry. We achieved this objective by replacing the pronoun with either the correct or the incorrect referent, thereby evaluating the  the discourse relation between them. The training process involved optimising two hyper-parameters: the regularisation parameter $c$ and the choice of kernel type, which could be either linear or a radial basis function (RBF). We leveraged a grid search technique with a 10 fold cross-validation scheme to identify the most suitable combination of hyper-parameters. The resulting SVM model with the best-tuned hyper-parameters was used for evaluation on the testing dataset. The results in Table \ref{f1score_resSVM} show that \textbf{SVM Add} achieved a lower F1 score of 0.821 in comparison to \textbf{SVM Full}, which achieved a solid F1 score of 0.914.
 





\begin{table}[h]
\centering
\begin{tabular}{lllll}
\hline
\textbf{Model} & \textbf{F1 Score}\\
\hline
\textbf{SVM Full} & 0.914 \\
\textbf{SVM Add} &0.821 \\
\hline
\end{tabular}
\caption{Evaluation performance of classical compositional and non-compositional SVM models}
\label{f1score_resSVM}
\end{table}


%


Additionally, we evaluated CoreNLP \cite{manning-etal-2014-stanford}, Neural Coreference \cite{clark-manning-2016-deep} \cite{clark-manning-2016-improving}, and SpanBERT \cite{lee2018}. CoreNLP combines rule-based techniques with statistical models to resolve coreference; Neural Coreference employs deep learning  to capture patterns and dependencies in text, and SpanBERT is a specialised version of  BERT  \cite{devlin2019bert}  fine-tuned for coreference resolution. We ran the pre-trained models using Stanza\footnote{https://corenlp.run/}, HuggingFace\footnote{https://huggingface.co/coref/}, and AllenNLP\footnote{https://demo.allennlp.org/coreference-resolution/} libraries respectively. The outcomes are presented in Table \ref{f1score_res}. 

\begin{table}[h]
\centering
\begin{tabular}{lllll}
\hline
\textbf{Model} & \textbf{F1 Score}\\
\hline
\textbf{CoreNLP} & 0.563 \\
\textbf{Neural Coreference} & 0.585 \\
\textbf{SpanBERT} & \textbf{0.927} \\
\hline
\textbf{QuantumCoref} & 0.872 \\
\hline
\end{tabular}
\caption{Evaluation performance of classical neural models}
\label{f1score_res}
\end{table}

The performance levels amongst these systems were diverse. CoreNLP achieved the lowest F1 score of 0.563, while SpanBERT demonstrated the highest score of 0.927. Neural Coreference achieved a moderate score of 0.585, trailing behind SpanBERT but outperforming CoreNLP. 

To facilitate the use of our approach, we implemented an end-to-end system named \textit{QuantumCoref} that consists of two sub-modules: (a) a mentions-detection module that uses SpaCy’s\footnote{https://spacy.io/usage/linguistic-features} part-of-speech parser to identify a set of potential coreference mentions, and (b) our highest-accurate trained $\mathbf{SLLM}$  classifier, which computes coreference scores for each pair of potential mentions. It achieved an F1 score of 0.872 near SpanBERT.

\begin{table*}[h]
\centering
\begin{tabular}{lllll}
\hline
\textbf{Model} & \textbf{F1 Score}\\
\hline
\textbf{CoreNLP + QuantumCoref} & 0.930 \\
\textbf{Neural Coreference + QuantumCoref} & 0.946\\
\textbf{SpanBERT+ QuantumCoref} & \textbf{0.986}\\
\hline
\textbf{SVM Full + QuantumCoref} & 0.959\\
\textbf{SVM Add + QuantumCoref} & 0.910\\
\hline
\end{tabular}
\caption{Evaluation performance of mixed quantum + classical models}
\label{f1score_resQuantumCoref}
\end{table*}

\subsection{Mixed Quantum + Classical Models}

To maximize the strengths of  quantum and classical systems, we combine their predictions in the following manner: when a classical system predicts an incorrect referent, we opt for the prediction of \textit{QuantumCoref}. Similarly, when a classical model fails to identify a referent, resulting in an empty cluster, we rely on \textit{QuantumCoref} for classification. As an example, consider the discourse \textit{``The students learned from the books. They were filled with knowledge.''} In this scenario, while SpanBERT detected that the pronoun \textit{``they''} refers to \textit{``students''}, \textit{QuantumCoref} correctly identified the coreference relationship as \textit{``they-books''}. As a result, this mixed quantum-classical approach recognised \textit{``they''} and \textit{``books''} as co-referent entities.  By combining the two approaches, we were able to extract the best outcomes from each model, thus enhancing the overall performance. CoreNLP improved from 0.563 to 0.930, Neural Coreference from 0.585 to 0.946, and SpanBERT from 0.927 to 0.986. The SVM models reacted in a similar fashion: the performance of SVM Add increased from  0.821 to 0.910 and that of SVM Full from 0.914 to 0.959.

\subsection{Discussion}
In a more detailed analysis, among the incorrect predictions, SpanBERT identified pronouns referring to the first noun in 95\% of the cases and to the second noun in 5\% of the cases. This highlights how SpanBERT struggles in identifying the correct referent, particularly when it's positioned towards the end of the sentence, leading to a higher preference for selecting the first noun. 

In situations characterised by linguistic ambiguities, SpanBERT struggles in recognising referential connections. Notably, in instances where multiple plausible nouns could serve as antecedents for pronouns, SpanBERT returns an empty cluster. For instance, in \textit{``The productive bee flew over the flower. It was magnificent.''} the complexity arises from the fact that both \textit{``productive bee''} and \textit{``flower''} are reasonable candidates for the antecedent. Similarly, in \textit{``The sailors jumped from the boats. They were having technical problems.''}, the ambiguity arises from the potential referents for the pronoun \textit{``They''} which could be either the \textit{``sailors''} or the \textit{``boats''}.
In contrast, \textit{QuantumCoref} relies on sentence structure and the connections between entities and their referents.  Impressively, \textit{QuantumCoref} solves 319 examples where SpanBERT misclassified, showcasing a success rate of 81.37\% and handled 35 examples where SpanBERT returned empty clusters, with a success rate of 68.62\%. When our dataset was converted into CoNLL format and SpanBERT was fine-tuned  on it, unsurprisingly, it achieved an F1 score of 0.998.

We would like to emphasise that these experiments were not specifically aimed at showcasing \textit{quantum advantage} over classical coreference resolution systems. Our aim  was to demonstrate the capabilities of our quantum-based approach, which also offers transparency. Furthermore, SpanBERT, with its exceptional coreference resolution capabilities, requires high computational resources. The fine-tuned SpanBERT model comprises a total of 366 million parameters, which is substantially larger compared to \textit{QuantumCoref}, with a total of 2693 parameters. This highlights the efficiency of the quantum-based approach. There is potential for further improvements, especially when a greater number of qubits are used in modelling. Our setting can  resolve general coreference relations in the same way as anaphoric ones. When multiple expressions co-refer, the main entity becomes a Fock space  and the rest are  pronoun types. We leave experimentation in this direction to future work.

\section*{Limitations}
We classify the limitations into the following  items:
\begin{itemize}
\item {\bf Syntax.} It would be tempting to call $\sllm$, the logic of discourse. It, however, does not have a connective for conjoining sentences. In this paper, we resolved the problem in the semantics, by using the Frobenius multiplication for conjoining sentences. A better logic for discourse should include this connective in its syntax. 
\item {\bf Semantics.} The vector space semantics of $\sllm$ over unifies the types, e.g. its copiable and functional types are assigned the same vector space semantics, e.g. two copies of a noun phrase and an adjective both have the same  $\semantics{N \otimes N}$ semantics. 
\item {\bf Automated Parsing.} $\sllm$ does not have an automatic parser and at the moment its use implies manual type annotations to words. LC has an automatic parser that can be extended to the new types introduced in $\sllm$. An automatic learning procedure  for types, however, requires a corpus annotated with $\sllm$ types. At this stage, we foresee any co-reference annotated corpus can easily be transferred to an $\sllm$ annotated one. 
\item {\bf Quantum Computation.} We relied  on simulations  for training circuit parameters instead of  using real quantum computers. Currently, we are experimenting with a shot-based simulation with an incorporated noise model. This approach takes into consideration critical factors such as quantum gate errors, decoherence, and shot noise, all of which affect practical quantum computing. It can be ported for execution on a quantum computer.

\item {\bf Different Types of Anaphora.}
In this paper, we focused on definite pronoun resolution and identity anaphora.   Non-definite and non-identity anaphora cases, such as bridging and event anaphora, pose challenges and require further theoretical work.

\item {\bf OntoNotes.}
Our original goal was to run the model on OntoNotes. This turned out to be impossible due to two main reasons. One was that we needed a large overlap between the vocabularies used in training and testing. Secondly, the entries of OntoNotes consist of long complex sentences, which would lead to large quantum circuits. These could not even be efficiently simulated with the current technology.

\section*{Acknowledgement}

The authors gratefully acknowledge the three anonymous reviewers for their valuable comments. Mehrnoosh Sadrzadeh is grateful to the Royal Academy of Engineering Research Chair/Senior Research Fellowship RCSRF2122-14-152 on Engineered Mathematics for Modelling Typed Structures. Hadi Wazni would like to express gratitude for support by UCL CS department for the PhD scholarships. Both authors would like to thank Lo Ian Kin for many helpful discussions and some help in implmentation.


\end{itemize}


\bibliography{anthology}
\bibliographystyle{acl_natbib}

\appendix

\newpage
\onecolumn
\section*{Appendix}
\label{sec:appendix}

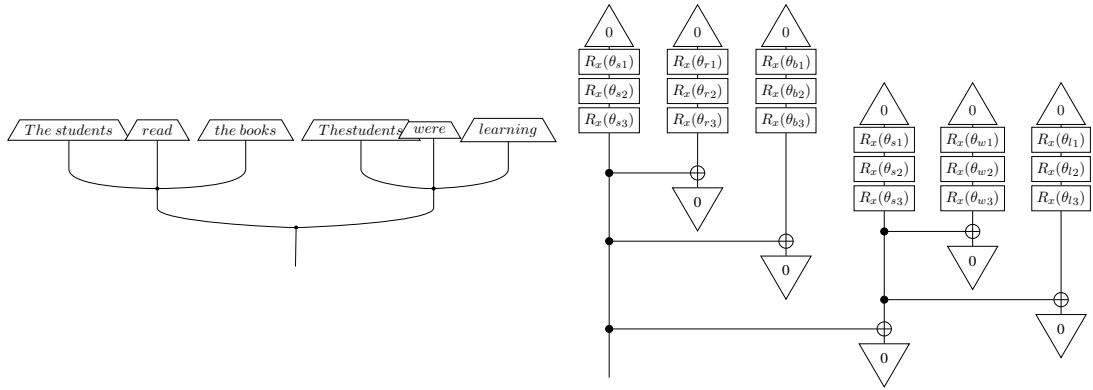
\begin{figure*}[h]
\centering
\resizebox{0.9\textwidth}{!}{%
      {%
\beginpgfgraphicnamed{example4_a}
\InputIfFileExists{example4_a.tikz}{}{\input{./tikz/example4_a.tikz}}
\endpgfgraphicnamed}
    }%
\caption{A Bag-of-Words diagram representing the discourse: \textit{``The students read the books. They were learninig.''} The diagram along with its transformation into a PQC.}
\end{figure*}

\begin{itemize}
\item Template Example 1:
\begin{itemize}
\item The sailors \textit{\{verb, phrasal verb, verb phrase\}} the boats. They were \textit{\{adjective, gerund phrase\}}
\item The \textit{\{adjective\}} sailors \textit{\{verb, phrasal verb, verb phrase\}} the boats. They were \textit{\{adjective, gerund phrase\}}
\item The sailors \textit{\{verb, phrasal verb, verb phrase\}} the \textit{\{adjective\}} boats. They were \textit{\{adjective, gerund phrase\}}
\item The \textit{\{adjective\}} sailors \textit{\{verb, phrasal verb, verb phrase\}} the \textit{\{adjective\}} boats. They were \textit{\{adjective, gerund phrase\}}
\end{itemize}
\end{itemize}

\begin{itemize}
\item Template Example 2:
\begin{itemize}
\item The storm \textit{\{verb, verb phrase\}} the flight. It was \textit{\{gerund phrase\}}
\item The \textit{\{adjective\}} storm \textit{\{verb, verb phrase\}} the flight. It was \textit{\{gerund phrase\}}
\item The storm \textit{\{verb, verb phrase\}} the \textit{\{adjective\}} flight. It was \textit{\{gerund phrase\}}
\item The \textit{\{adjective\}} storm \textit{\{verb, verb phrase\}} the \textit{\{adjective\}} flight. It was \textit{\{gerund phrase\}}
\end{itemize}
\item 8 distinct structural patterns for Template Example 2:
\begin{itemize}
\item The storm \textit{\{verb, verb phrase\}} the flight. It was \textit{\{gerund phrase (storm)\}}.
\item The storm \textit{\{verb, verb phrase\}} the flight. It was \textit{\{gerund phrase (flight)\}}.
\item The \textit{\{adjective (storm)\}} storm \textit{\{verb, verb phrase\}} the flight. It was \textit{\{gerund phrase (storm)\}}.
\item The \textit{\{adjective (storm)\}} storm \textit{\{verb, verb phrase\}} the flight. It was \textit{\{gerund phrase (flight)\}}.
\item The storm \textit{\{verb, verb phrase\}} the \textit{\{adjective (flight)\}} flight. It was \textit{\{gerund phrase (flight)\}}.
\item The storm \textit{\{verb, verb phrase\}} the \textit{\{adjective (flight)\}} flight. It was \textit{\{gerund phrase (storm)\}}.
\item The \textit{\{adjective (storm)\}} storm \textit{\{verb, verb phrase\}} the \textit{\{adjective (flight)\}} flight. It was \textit{\{gerund phrase (storm)\}}.
\item The \textit{\{adjective (storm)\}} storm \textit{\{verb, verb phrase\}} the \textit{\{adjective (flight)\}} flight. It was \textit{\{gerund phrase (flight)\}}.
\end{itemize}
\end{itemize}

\end{document}

%% file: tikzpreamble.tex
\usepackage{tikz}
\usepackage{pgfplots}
\usetikzlibrary{trees}
\usetikzlibrary{topaths}
\usetikzlibrary{decorations.pathmorphing}
\usetikzlibrary{decorations.markings}
\usetikzlibrary{matrix,backgrounds,folding}
\usetikzlibrary{chains,scopes,positioning,fit}
\usetikzlibrary{shapes,shapes.geometric,shapes.misc}

\tikzset{->-/.style={decoration={
  markings,
  mark=at position .5 with {\arrow{>}}},postaction={decorate}}}

\pgfdeclarelayer{edgelayer}
\pgfdeclarelayer{nodelayer}
\pgfsetlayers{background,edgelayer,nodelayer,main}

\tikzstyle{none}=[fill=none]
\tikzstyle{title}=[fill=none]
\tikzstyle{lab}=[fill=white, font=\tiny]
\tikzstyle{new style 0}=[fill=white, draw=black, shape=trapezium]
\tikzstyle{new style 1}=[fill=white, draw=black, shape=circle]
\tikzstyle{new style 2}=[fill=white, draw=green, shape=circle]
\tikzstyle{new style 3}=[fill=white, draw=red, shape=circle]
\tikzstyle{new style 4}=[fill=white, draw=black, thick, shape=regular polygon, regular polygon sides=3]
\tikzstyle{element}=[fill=white, draw=black, shape=trapezium]
\tikzstyle{functional}=[fill=white, draw=black, shape=trapezium,shape border rotate=180]
\tikzstyle{counit}=[fill=black, shape=circle]
\tikzstyle{bigClasp}=[fill=white, draw=black, shape=circle, minimum width=1.7cm]

\tikzstyle{downArrow}=[none]
\tikzstyle{thickArr}=[line width = 2pt]
\tikzstyle{thick}=[line width = 2pt]
\tikzstyle{thickerArr}=[->-, line width = 4pt]
\tikzstyle{thicker}=[line width = 4pt]
\tikzstyle{dashed}=[-, dashed]
\tikzstyle{every loop}=[]

\input{ian.tikzstyles}
\tikzstyle{(null)}=[]
\tikzstyle{plain}=[]
\tikzstyle{simple}=[]

%% file: ian.tikzstyles
\tikzstyle{state}=[fill=white, draw=black, shape=regular polygon, regular polygon sides=3, scale=1]
\tikzstyle{effect}=[fill=white, draw=black, shape=regular polygon, regular polygon sides=3, shape border rotate=180, scale=1]
\tikzstyle{control}=[fill=black, draw=black, shape=circle, scale=0.5]
\tikzstyle{plus}=[fill=white, draw=black, shape=circle, append after command={(\tikzlastnode.north) edge (\tikzlastnode.south) (\tikzlastnode.east) edge (\tikzlastnode.west)}, scale=1]
\tikzstyle{process}=[fill=white, draw=black, shape=rectangle]
\tikzstyle{simple}=[]

\tikzstyle{1storage}=[fill=white, draw=black, shape=trapezium, scale=1,minimum height=15, minimum width=30, trapezium stretches body, trapezium angle=45]
\tikzstyle{2storage}=[fill=white, draw=black, shape=trapezium, scale=1,minimum height=15, minimum width=50, trapezium stretches body, trapezium angle=45]
\tikzstyle{3storage}=[fill=white, draw=black, shape=trapezium, scale=1,minimum height=15, minimum width=70, trapezium stretches body, trapezium angle=45]

\tikzstyle{1costorage}=[fill=white, draw=black, shape=trapezium, scale=1,minimum height=15, minimum width=30, trapezium stretches body, trapezium angle=45, shape border rotate=180]
\tikzstyle{2costorage}=[fill=white, draw=black, shape=trapezium, scale=1,minimum height=15, minimum width=50, trapezium stretches body, trapezium angle=45, shape border rotate=180]
\tikzstyle{3costorage}=[fill=white, draw=black, shape=trapezium, scale=1,minimum height=15, minimum width=70, trapezium stretches body, trapezium angle=45, shape border rotate=180]

\tikzstyle{regular}=[-]

%% file: tikz/translation.tikz
\begin{tikzpicture}
	\begin{pgfonlayer}{nodelayer}
		\node [style=process] (4) at (12.5, -15.35) {$H$};
		\node [style=control] (7) at (12.5, -14.6) {};
		\node [style=none] (9) at (12.5, -14.1) {};
		\node [style=none] (10) at (13.75, -14.1) {};
		\node [style=effect] (17) at (12.5, -16.1) {0};
		\node [style=effect] (18) at (13.75, -16.1) {0};
		\node [style=plus] (19) at (13.75, -14.6) {};
		\node [style=none] (21) at (8.5, -15.25) {};
		\node [style=none] (22) at (9.75, -15.25) {};
		\node [style=none] (23) at (11, -15.5) {$\mapsto$};
		\node [style=process] (43) at (5.75, -14.75) {$H$};
		\node [style=process] (44) at (4.5, -14.75) {$H$};
		\node [style=process] (45) at (7, -14.75) {$H$};
		\node [style=control] (46) at (4.5, -15.5) {};
		\node [style=control] (47) at (5.75, -16.25) {};
		\node [style=process] (48) at (5.75, -15.5) {$R_z(\theta_{01})$};
		\node [style=process] (49) at (7, -16.25) {$R_z(\theta_{12})$};
		\node [style=none] (50) at (7, -16.75) {};
		\node [style=none] (51) at (5.75, -16.75) {};
		\node [style=none] (52) at (4.5, -16.75) {};
		\node [style=state] (53) at (4.5, -14) {$0$};
		\node [style=state] (54) at (5.75, -14) {$0$};
		\node [style=state] (55) at (7, -14) {$0$};
		\node [style=state] (82) at (-8.5, -14) {$0$};
		\node [style=process] (83) at (-8.5, -14.75) {$R_x(\theta_1)$};
		\node [style=process] (84) at (-8.5, -15.5) {$R_z(\theta_2)$};
		\node [style=process] (85) at (-8.5, -16.25) {$R_x(\theta_3)$};
		\node [style=none] (86) at (-8.5, -16.75) {};
		\node [style=none] (112) at (0, -16) {};
		\node [style=none] (113) at (0.75, -16) {};
		\node [style=none] (114) at (1.5, -16) {};
		\node [style=none] (115) at (0, -15.25) {};
		\node [style=none] (116) at (0.75, -15.25) {};
		\node [style=none] (117) at (1.5, -15.25) {};
		\node [style=none] (121) at (3, -15.5) {$\mapsto$};
		\node [style=none] (122) at (-10.25, -15.5) {$\mapsto$};
		\node [style=none] (123) at (-11.75, -16) {};
		\node [style=none] (124) at (-11.75, -15.25) {};
		\node [style=1storage] (126) at (-11.75, -15.25) {};
		\node [style=3storage] (127) at (0.75, -15.25) {};
		\node [style=process] (128) at (-2, -14.75) {$H$};
		\node [style=process] (129) at (-3.25, -14.75) {$H$};
		\node [style=control] (131) at (-3.25, -15.5) {};
		\node [style=process] (133) at (-2, -15.5) {$R_z(\theta_{01})$};
		\node [style=none] (136) at (-2, -16.75) {};
		\node [style=none] (137) at (-3.25, -16.75) {};
		\node [style=state] (138) at (-3.25, -14) {$0$};
		\node [style=state] (139) at (-2, -14) {$0$};
		\node [style=none] (141) at (-6.75, -16) {};
		\node [style=none] (142) at (-5.75, -16) {};
		\node [style=none] (144) at (-6.75, -15.25) {};
		\node [style=none] (145) at (-5.75, -15.25) {};
		\node [style=none] (147) at (-4.5, -15.5) {$\mapsto$};
		\node [style=2storage] (148) at (-6.25, -15.25) {};
		\node [style=none] (209) at (17.25, -15.6) {$\mapsto$};
		\node [style=control] (210) at (18.75, -15.25) {};
		\node [style=none] (211) at (18.75, -14.5) {};
		\node [style=none] (212) at (20, -14.5) {};
		\node [style=plus] (213) at (20, -15.25) {};
		\node [style=none] (214) at (18.75, -16.1) {};
		\node [style=effect] (215) at (20, -16.1) {0};
		\node [style=control] (216) at (15.5, -15.6) {};
		\node [style=none] (217) at (15, -15.1) {};
		\node [style=none] (218) at (16, -15.1) {};
		\node [style=none] (219) at (15.5, -16.1) {};
	\end{pgfonlayer}
	\begin{pgfonlayer}{edgelayer}
		\draw [style=simple] (4) to (9.center);
		\draw [style=simple] (17) to (4);
		\draw [style=simple] (7) to (19);
		\draw [style=simple] (18) to (10.center);
		\draw [style=regular, bend right=90, looseness=1.50] (21.center) to (22.center);
		\draw [style=simple] (46) to (48);
		\draw [style=simple] (47) to (49);
		\draw [style=regular] (53) to (52.center);
		\draw [style=regular] (54) to (51.center);
		\draw [style=regular] (55) to (50.center);
		\draw [style=regular] (82) to (86.center);
		\draw [style=regular] (115.center) to (112.center);
		\draw [style=regular] (116.center) to (113.center);
		\draw [style=regular] (117.center) to (114.center);
		\draw [style=regular] (124.center) to (123.center);
		\draw [style=simple] (131) to (133);
		\draw [style=regular] (138) to (137.center);
		\draw [style=regular] (139) to (136.center);
		\draw [style=regular] (144.center) to (141.center);
		\draw [style=regular] (145.center) to (142.center);
		\draw [style=simple] (210) to (213);
		\draw [style=simple] (215) to (212.center);
		\draw [style=regular, bend right, looseness=0.75] (216) to (218.center);
		\draw [style=regular, bend right, looseness=0.75] (217.center) to (216);
		\draw [style=regular] (216) to (219.center);
		\draw [style=regular] (214.center) to (211.center);
	\end{pgfonlayer}
\end{tikzpicture}

%% file: tikz/example1_c.tikz
\begin{tikzpicture}[baseline={(0,0)}]
	\begin{pgfonlayer}{nodelayer}
		\node [style=none] (261) at (2.1833, -1.25) {};
		\node [style=none] (262) at (2.1833, -2.75) {};
		\node [style=none, fill=white, right] (263) at (1.5333, -1.6) {$N$};
		\node [style=none] (264) at (6.1833, -1.25) {};
		\node [style=none] (265) at (6.1833, -4.25) {};
		\node [style=none, fill=white, right] (266) at (6.2833, -1.6) {$N$};
		\node [style=none] (267) at (3.1833, -2.25) {};
		\node [style=none] (268) at (3.1833, -2.75) {};
		\node [style=none, fill=white, right] (269) at (2.5333, -2.6) {$N$};
		\node [style=none] (270) at (3.4333, -2.25) {};
		\node [style=none] (271) at (3.4333, -6) {};
		\node [style=none] (272) at (3.6833, -2.25) {};
		\node [style=none] (273) at (3.6833, -4) {};
		\node [style=none, fill=white, right] (274) at (3.7833, -2.6) {$N$};
		\node [style=none] (276) at (4.9333, -3.5) {};
		\node [style=none] (277) at (4.9333, -4) {};
		\node [style=none, fill=white, right] (278) at (5.0333, -3.85) {$N$};
		\node [style=none, fill=white, right] (282) at (7.2833, -4.1) {$N$};
		\node [style=none] (284) at (7.1833, -5.25) {};
		\node [style=none, fill=white, right] (285) at (6.2833, -4.1) {$N$};
		\node [style=none] (287) at (8.1833, -4.5) {};
		\node [style=none] (288) at (8.1833, -5.25) {};
		\node [style=none, fill=white, right] (289) at (7.5333, -5.1) {$N$};
		\node [style=none] (290) at (8.4333, -4.5) {};
		\node [style=none] (291) at (8.4333, -6) {};
		\node [style=none] (292) at (8.6833, -4.5) {};
		\node [style=none, fill=white, right] (294) at (8.7833, -5.1) {$N$};
		\node [style=none] (296) at (5.9333, -6.25) {};
		\node [style=none] (297) at (5.9333, -6.5) {};
		\node [style=none] (298) at (5.9333, -7.75) {};
		\node [style=none, fill=white, right] (299) at (6.2833, -7.35) {$S$};
		\node [circle, fill=black, scale=0.309] (300) at (5.9333, -6.25) {};
		\node [style=none] (307) at (3.9333, -2.25) {};
		\node [style=element] (331) at (3.9333, -1) {$ ~~~~~~~~~~~The~students ~~~~~~~~~~~~~~~~$};
		\node [style=element] (332) at (3.6833, -2) {$read$};
		\node [style=element] (333) at (4.9333, -3.25) {$the~books$};
		\node [style=element] (334) at (7.1833, -3.5) {$They$};
		\node [style=none] (335) at (6.9333, -3.5) {};
		\node [style=none] (336) at (6.9333, -4.25) {};
		\node [style=none] (337) at (7.1833, -3.5) {};
		\node [style=none] (338) at (7.1833, -5.25) {};
		\node [style=element] (339) at (8.4333, -4.5) {$were$};
		\node [style=none] (365) at (7.25, -6.5) {};
		\node [style=none] (366) at (7.5, -7) {};
		\node [style=none] (367) at (9.75, -7) {};
		\node [style=none] (368) at (10, -6.5) {};
		\node [style=none] (369) at (8.75, -6.75) {$learning$};
		\node [style=none] (370) at (8.75, -4.5) {};
		\node [style=none] (371) at (8.75, -6.5) {};
	\end{pgfonlayer}
	\begin{pgfonlayer}{edgelayer}
		\draw [in=90, out=-90] (261.center) to (262.center);
		\draw [in=90, out=-90] (264.center) to (265.center);
		\draw [in=90, out=-90] (267.center) to (268.center);
		\draw [in=90, out=-90] (270.center) to (271.center);
		\draw [in=90, out=-90] (272.center) to (273.center);
		\draw [in=90, out=-90] (276.center) to (277.center);
		\draw [in=90, out=-90] (287.center) to (288.center);
		\draw [in=90, out=-90] (290.center) to (291.center);
		\draw [in=90, out=-90] (296.center) to (297.center);
		\draw [in=180, out=-90, looseness=0.15] (271.center) to (296.center);
		\draw [in=0, out=-90, looseness=0.15] (291.center) to (296.center);
		\draw [in=90, out=-90] (297.center) to (298.center);
		\draw [in=-90, out=-90, looseness=1.25] (262.center) to (268.center);
		\draw [in=-90, out=-90, looseness=0.75] (273.center) to (277.center);
		\draw [in=90, out=-90] (335.center) to (336.center);
		\draw [in=90, out=-90] (337.center) to (338.center);
		\draw [in=-90, out=-90] (265.center) to (336.center);
		\draw [bend right=90] (338.center) to (288.center);
		\draw (365.center) to (368.center);
		\draw (368.center) to (367.center);
		\draw (367.center) to (366.center);
		\draw (366.center) to (365.center);
		\draw (370.center) to (371.center);
	\end{pgfonlayer}
\end{tikzpicture}

%% file: tikz/example1_d.tikz
\begin{tikzpicture}[baseline={(0,0)}]
	\begin{pgfonlayer}{nodelayer}
		\node [style=process] (372) at (2.6833, -4.5) {$H$};
		\node [style=process] (373) at (1.4333, -1.75) {$H$};
		\node [style=process] (374) at (3.9333, -4.5) {$H$};
		\node [style=control] (375) at (1.4333, -2.5) {};
		\node [style=control] (376) at (2.6833, -5.25) {};
		\node [style=process] (378) at (3.9333, -5.25) {$R_z({\theta_r}_{01})$};
		\node [style=none] (380) at (3.9333, -12.25) {};
		\node [style=state] (381) at (1.4333, -1) {$0$};
		\node [style=state] (382) at (2.6833, -3.75) {$0$};
		\node [style=state] (383) at (3.9333, -3.75) {$0$};
		\node [style=effect] (389) at (7.1833, -10.1) {0};
		\node [style=plus] (390) at (7.1833, -8.6) {};
		\node [style=process] (391) at (8.6833, -1.75) {$H$};
		\node [style=process] (394) at (8.6833, -2.5) {$R_z({\theta_s}_{01})$};
		\node [style=state] (396) at (8.6833, -1) {$0$};
		\node [style=process] (397) at (8.6833, -6.1) {$H$};
		\node [style=control] (398) at (8.6833, -5.35) {};
		\node [style=effect] (399) at (8.6833, -6.85) {0};
		\node [style=effect] (400) at (10.1833, -6.85) {0};
		\node [style=plus] (401) at (10.1833, -5.35) {};
		\node [style=process] (402) at (11.6833, -4) {$H$};
		\node [style=process] (403) at (10.1833, -4) {$H$};
		\node [style=state] (404) at (10.1833, -3.25) {$0$};
		\node [style=state] (405) at (11.6833, -3.25) {$0$};
		\node [style=process] (406) at (11.6833, -7.35) {$H$};
		\node [style=control] (407) at (11.6833, -6.6) {};
		\node [style=effect] (408) at (11.6833, -8.1) {0};
		\node [style=effect] (409) at (13.1833, -8.1) {0};
		\node [style=plus] (410) at (13.1833, -6.6) {};
		\node [style=process] (411) at (14.6833, -5.25) {$H$};
		\node [style=process] (412) at (13.1833, -5.25) {$H$};
		\node [style=process] (413) at (16.1833, -5.25) {$H$};
		\node [style=control] (414) at (13.1833, -6) {};
		\node [style=control] (415) at (14.6833, -6.75) {};
		\node [style=process] (416) at (14.6833, -6) {$R_z({\theta_w}_{01})$};
		\node [style=process] (417) at (16.1833, -6.75) {$R_z({\theta_w}_{12})$};
		\node [style=state] (418) at (13.1833, -4.5) {$0$};
		\node [style=state] (419) at (14.6833, -4.5) {$0$};
		\node [style=state] (420) at (16.1833, -4.5) {$0$};
		\node [style=effect] (421) at (16.1833, -12) {$0$};
		\node [style=process] (422) at (16.1833, -11.25) {$R_x(-{\theta_l}_1)$};
		\node [style=process] (423) at (16.1833, -10.5) {$R_x(-{\theta_l}_2)$};
		\node [style=process] (424) at (16.1833, -9.75) {$R_x(-{\theta_l}_3)$};
		\node [style=plus] (425) at (14.6833, -11.1) {};
		\node [style=control] (426) at (3.9333, -11.1) {};
		\node [style=plus] (427) at (14.6833, -11.1) {};
		\node [style=none] (428) at (10.1833, -4.85) {};
		\node [style=control] (429) at (10.1833, -4.75) {};
		\node [style=process] (430) at (11.6833, -4.75) {$R_z({\theta_t}_{01})$};
		\node [style=effect] (431) at (14.6833, -12) {$0$};
		\node [style=process] (437) at (1.4333, -7.6) {$H$};
		\node [style=control] (438) at (1.4333, -6.85) {};
		\node [style=effect] (439) at (1.4333, -8.35) {0};
		\node [style=plus] (441) at (2.6833, -6.85) {};
		\node [style=process] (442) at (5.1833, -9.35) {$H$};
		\node [style=control] (443) at (5.1833, -8.6) {};
		\node [style=effect] (444) at (5.1833, -10.1) {0};
		\node [style=process] (445) at (5.1833, -4.5) {$H$};
		\node [style=state] (447) at (5.1833, -3.75) {$0$};
		\node [style=process] (448) at (5.1833, -6) {$R_z({\theta_r}_{12})$};
		\node [style=control] (449) at (3.9333, -6) {};
		\node [style=process] (450) at (2.6833, -7.6) {$H$};
		\node [style=effect] (451) at (2.6833, -8.35) {0};
		\node [style=process] (452) at (7.1833, -6.25) {$R_x({\theta_b}_1)$};
		\node [style=process] (453) at (7.1833, -7) {$R_x({\theta_b}_2)$};
		\node [style=process] (454) at (7.1833, -7.75) {$R_x({\theta_b}_3)$};
		\node [style=process] (456) at (7.1833, -5.5) {$H$};
		\node [style=state] (457) at (7.1833, -4.75) {$0$};
	\end{pgfonlayer}
	\begin{pgfonlayer}{edgelayer}
		\draw [style=simple] (376) to (378);
		\draw (396) to (391);
		\draw (391) to (394);
		\draw [style=simple] (399) to (397);
		\draw [style=simple] (398) to (401);
		\draw (394) to (398);
		\draw (398) to (397);
		\draw (404) to (403);
		\draw (405) to (402);
		\draw [style=simple] (408) to (406);
		\draw [style=simple] (407) to (410);
		\draw (407) to (406);
		\draw [style=simple] (414) to (416);
		\draw [style=simple] (415) to (417);
		\draw (419) to (411);
		\draw (411) to (416);
		\draw (416) to (415);
		\draw (418) to (412);
		\draw (412) to (414);
		\draw (414) to (410);
		\draw (415) to (425);
		\draw [style=simple] (426) to (427);
		\draw [style=simple] (429) to (430);
		\draw (403) to (429);
		\draw (429) to (401);
		\draw (401) to (400);
		\draw (402) to (430);
		\draw (430) to (407);
		\draw (410) to (409);
		\draw (427) to (431);
		\draw (421) to (422);
		\draw (422) to (423);
		\draw (423) to (424);
		\draw (424) to (417);
		\draw (417) to (413);
		\draw (413) to (420);
		\draw (381) to (373);
		\draw (373) to (375);
		\draw [style=simple] (439) to (437);
		\draw [style=simple] (438) to (441);
		\draw (375) to (438);
		\draw (438) to (437);
		\draw (390) to (389);
		\draw [style=simple] (444) to (442);
		\draw (383) to (374);
		\draw (374) to (378);
		\draw (449) to (448);
		\draw (447) to (445);
		\draw (445) to (448);
		\draw [style=simple] (451) to (450);
		\draw (441) to (450);
		\draw (426) to (380.center);
		\draw (448) to (443);
		\draw (443) to (442);
		\draw (382) to (441);
		\draw (378) to (449);
		\draw (449) to (426);
		\draw (443) to (390);
		\draw (457) to (454);
		\draw (454) to (390);
		\draw (375) to (394);
	\end{pgfonlayer}
\end{tikzpicture}

%% file: tikz/example2_c.tikz
\begin{tikzpicture}[baseline={(0,0)}]
	\begin{pgfonlayer}{nodelayer}
		\node [style=none] (371) at (2, -1) {};
		\node [style=none, fill=white, right] (373) at (1.1, -1.35) {$N$};
		\node [style=none] (374) at (3, -1) {};
		\node [style=none] (375) at (3, -5.25) {};
		\node [style=none] (377) at (3.5, -1) {};
		\node [style=none] (378) at (3.5, -3) {};
		\node [style=none, fill=white, right] (379) at (3.6, -1.35) {$N$};
		\node [style=none] (380) at (4.25, -2.5) {};
		\node [style=none] (381) at (4.25, -3) {};
		\node [style=none, fill=white, right] (382) at (3.6, -2.85) {$N$};
		\node [style=none] (383) at (5, -2.5) {};
		\node [style=none] (384) at (5, -3.75) {};
		\node [style=none, fill=white, right] (385) at (5.1, -2.85) {$N$};
		\node [style=none] (387) at (6.25, -3.25) {};
		\node [style=none] (388) at (6.25, -3.75) {};
		\node [style=none, fill=white, right] (389) at (5.6, -3.6) {$N$};
		\node [style=none] (390) at (6.75, -3.25) {};
		\node [style=none] (391) at (6.75, -4.75) {};
		\node [style=none, fill=white, right] (392) at (6.85, -3.6) {$N$};
		\node [style=none] (393) at (5.75, -4) {};
		\node [style=none] (394) at (7.75, -4) {};
		\node [style=none] (395) at (7.75, -4.75) {};
		\node [style=none, fill=white, right] (396) at (7.1, -4.6) {$N$};
		\node [style=none] (397) at (8.25, -4) {};
		\node [style=none] (398) at (8.25, -5.25) {};
		\node [style=none] (400) at (8.75, -4) {};
		\node [style=none, fill=white, right] (402) at (8.85, -4.6) {$N$};
		\node [style=none] (403) at (7.25, -5) {};
		\node [style=none] (404) at (5.5, -5.5) {};
		\node [style=none] (405) at (5.5, -5.75) {};
		\node [style=none] (406) at (5.5, -7) {};
		\node [style=none, fill=white, right] (407) at (5.85, -6.6) {$S$};
		\node [circle, fill=black, scale=0.324] (408) at (5.5, -5.5) {};
		\node [style=none] (420) at (5.25, -2.5) {};
		\node [style=element] (439) at (2.75, -0.75) {$~~~~read~~~~$};
		\node [style=element] (440) at (4.75, -2.25) {$the~books$};
		\node [style=element] (441) at (6.5, -3) {$They$};
		\node [style=element] (442) at (8.25, -4) {$were$};
		\node [style=none] (443) at (-0.25, -1.75) {};
		\node [style=none] (444) at (0, -2.25) {};
		\node [style=none] (445) at (2.5, -2.25) {};
		\node [style=none] (446) at (2.75, -1.75) {};
		\node [style=none] (447) at (1.25, -2) {$The~students$};
		\node [style=none] (448) at (2, -1.75) {};
		\node [style=none] (449) at (7.5, -5.75) {};
		\node [style=none] (450) at (7.75, -6.25) {};
		\node [style=none] (451) at (9.75, -6.25) {};
		\node [style=none] (452) at (10, -5.75) {};
		\node [style=none] (453) at (8.75, -6) {$~~~~~interesting~~~~~$};
		\node [style=none] (454) at (8.75, -5.75) {};
	\end{pgfonlayer}
	\begin{pgfonlayer}{edgelayer}
		\draw [in=90, out=-90] (374.center) to (375.center);
		\draw [in=90, out=-90] (377.center) to (378.center);
		\draw [in=90, out=-90] (380.center) to (381.center);
		\draw [in=90, out=-90] (383.center) to (384.center);
		\draw [in=90, out=-90] (387.center) to (388.center);
		\draw [in=90, out=-90] (390.center) to (391.center);
		\draw [in=180, out=-90, looseness=0.94] (384.center) to (393.center);
		\draw [in=0, out=-90, looseness=0.94] (388.center) to (393.center);
		\draw [in=90, out=-90] (394.center) to (395.center);
		\draw [in=90, out=-90] (397.center) to (398.center);
		\draw [in=180, out=-90, looseness=0.94] (391.center) to (403.center);
		\draw [in=0, out=-90, looseness=0.94] (395.center) to (403.center);
		\draw [in=90, out=-90] (404.center) to (405.center);
		\draw [in=180, out=-90, looseness=0.15] (375.center) to (404.center);
		\draw [in=0, out=-90, looseness=0.15] (398.center) to (404.center);
		\draw [in=90, out=-90] (405.center) to (406.center);
		\draw (443.center) to (446.center);
		\draw (446.center) to (445.center);
		\draw (445.center) to (444.center);
		\draw (444.center) to (443.center);
		\draw (449.center) to (452.center);
		\draw (452.center) to (451.center);
		\draw (451.center) to (450.center);
		\draw (450.center) to (449.center);
		\draw (400.center) to (454.center);
		\draw (371.center) to (448.center);
		\draw [bend right=90] (378.center) to (381.center);
	\end{pgfonlayer}
\end{tikzpicture}

%% file: tikz/example2_d.tikz
\begin{tikzpicture}[baseline={(0,0)}]
	\begin{pgfonlayer}{nodelayer}
		\node [style=process] (490) at (2.75, -1.5) {$H$};
		\node [style=process] (491) at (1.25, -1.5) {$H$};
		\node [style=process] (492) at (4.25, -1.5) {$H$};
		\node [style=control] (493) at (1.25, -2.25) {};
		\node [style=control] (494) at (2.75, -3) {};
		\node [style=process] (495) at (2.75, -2.25) {$R_z({\theta_r}_{01})$};
		\node [style=process] (496) at (4.25, -3) {$R_z({\theta_r}_{12})$};
		\node [style=none] (497) at (4.25, -2.25) {};
		\node [style=none] (498) at (2.75, -10.25) {};
		\node [style=state] (500) at (1.25, -0.75) {$0$};
		\node [style=state] (501) at (2.75, -0.75) {$0$};
		\node [style=state] (502) at (4.25, -0.75) {$0$};
		\node [style=process] (503) at (4.25, -4.85) {$H$};
		\node [style=control] (504) at (4.25, -4.1) {};
		\node [style=none] (505) at (4.25, -2.1) {};
		\node [style=none] (506) at (5.75, -3.6) {};
		\node [style=effect] (507) at (4.25, -5.6) {0};
		\node [style=effect] (508) at (5.75, -5.6) {0};
		\node [style=plus] (509) at (5.75, -4.1) {};
		\node [style=process] (510) at (7.25, -2.75) {$H$};
		\node [style=process] (511) at (5.75, -2.75) {$H$};
		\node [style=control] (512) at (5.75, -3.5) {};
		\node [style=process] (513) at (7.25, -3.5) {$R_z({\theta_b}_{01})$};
		\node [style=state] (514) at (5.75, -2) {$0$};
		\node [style=state] (515) at (7.25, -2) {$0$};
		\node [style=process] (516) at (7.25, -5.85) {$H$};
		\node [style=control] (517) at (7.25, -5.1) {};
		\node [style=effect] (518) at (7.25, -6.6) {0};
		\node [style=effect] (519) at (8.75, -6.6) {0};
		\node [style=plus] (520) at (8.75, -5.1) {};
		\node [style=process] (522) at (10.25, -3.75) {$H$};
		\node [style=process] (523) at (8.75, -3.75) {$H$};
		\node [style=state] (526) at (8.75, -3) {$0$};
		\node [style=state] (527) at (10.25, -3) {$0$};
		\node [style=process] (528) at (10.25, -7.1) {$H$};
		\node [style=control] (529) at (10.25, -6.35) {};
		\node [style=effect] (530) at (10.25, -7.85) {0};
		\node [style=effect] (531) at (11.75, -7.85) {0};
		\node [style=plus] (532) at (11.75, -6.35) {};
		\node [style=process] (541) at (13.25, -5) {$H$};
		\node [style=process] (542) at (11.75, -5) {$H$};
		\node [style=process] (543) at (14.75, -5) {$H$};
		\node [style=control] (544) at (11.75, -5.75) {};
		\node [style=control] (545) at (13.25, -6.5) {};
		\node [style=process] (546) at (13.25, -5.75) {$R_z({\theta_w}_{01})$};
		\node [style=process] (547) at (14.75, -6.5) {$R_z({\theta_w}_{12})$};
		\node [style=state] (550) at (11.75, -4.25) {$0$};
		\node [style=state] (551) at (13.25, -4.25) {$0$};
		\node [style=state] (552) at (14.75, -4.25) {$0$};
		\node [style=effect] (554) at (14.75, -10) {$0$};
		\node [style=process] (555) at (14.75, -9.25) {$R_x(-{\theta_i}_1)$};
		\node [style=process] (556) at (14.75, -8.5) {$R_x(-{\theta_i}_2)$};
		\node [style=process] (557) at (14.75, -7.75) {$R_x(-{\theta_i}_3)$};
		\node [style=plus] (562) at (13.25, -9.1) {};
		\node [style=control] (564) at (2.75, -9.1) {};
		\node [style=plus] (565) at (13.25, -9.1) {};
		\node [style=none] (566) at (8.75, -4.6) {};
		\node [style=control] (567) at (8.75, -4.5) {};
		\node [style=process] (568) at (10.25, -4.5) {$R_z({\theta_t}_{01})$};
		\node [style=effect] (569) at (13.25, -10) {$0$};
		\node [style=effect] (570) at (1.25, -5.5) {$0$};
		\node [style=process] (571) at (1.25, -4.75) {$R_x(-{\theta_s}_1)$};
		\node [style=process] (572) at (1.25, -4) {$R_x(-{\theta_s}_2)$};
		\node [style=process] (573) at (1.25, -3.25) {$R_x(-{\theta_s}_3)$};
	\end{pgfonlayer}
	\begin{pgfonlayer}{edgelayer}
		\draw [style=simple] (493) to (495);
		\draw [style=simple] (494) to (496);
		\draw [style=regular] (501) to (498.center);
		\draw [style=regular] (502) to (497.center);
		\draw [style=simple] (503) to (505.center);
		\draw [style=simple] (507) to (503);
		\draw [style=simple] (504) to (509);
		\draw [style=simple] (508) to (506.center);
		\draw [style=simple] (512) to (513);
		\draw (514) to (511);
		\draw (511) to (512);
		\draw (515) to (510);
		\draw (510) to (513);
		\draw [style=simple] (518) to (516);
		\draw [style=simple] (517) to (520);
		\draw (513) to (517);
		\draw (517) to (516);
		\draw (526) to (523);
		\draw (527) to (522);
		\draw [style=simple] (530) to (528);
		\draw [style=simple] (529) to (532);
		\draw (529) to (528);
		\draw [style=simple] (544) to (546);
		\draw [style=simple] (545) to (547);
		\draw (551) to (541);
		\draw (541) to (546);
		\draw (546) to (545);
		\draw (550) to (542);
		\draw (542) to (544);
		\draw (544) to (532);
		\draw (545) to (562);
		\draw [style=simple] (564) to (565);
		\draw [style=simple] (567) to (568);
		\draw (523) to (567);
		\draw (567) to (520);
		\draw (520) to (519);
		\draw (522) to (568);
		\draw (568) to (529);
		\draw (532) to (531);
		\draw (565) to (569);
		\draw (554) to (555);
		\draw (555) to (556);
		\draw (556) to (557);
		\draw (557) to (547);
		\draw (547) to (543);
		\draw (543) to (552);
		\draw (570) to (571);
		\draw (571) to (572);
		\draw (572) to (573);
		\draw (500) to (491);
		\draw (491) to (493);
		\draw (493) to (573);
	\end{pgfonlayer}
\end{tikzpicture}

%% file: tikz/example4_a.tikz
\begin{tikzpicture}[baseline={(0,0)}]
	\begin{pgfonlayer}{nodelayer}
		\node [style=none] (944) at (0.91647, -3.5) {};
		\node [style=none] (945) at (0.91647, -4.5) {};
		\node [style=none] (947) at (3.16647, -3.5) {};
		\node [style=none] (948) at (3.16647, -4.75) {};
		\node [style=none] (950) at (5.41647, -3.5) {};
		\node [style=none] (951) at (5.41647, -4.5) {};
		\node [style=none] (955) at (3.16647, -5) {};
		\node [style=none] (956) at (3.16647, -5.25) {};
		\node [style=none] (957) at (3.16647, -5.5) {};
		\node [circle, fill=black, scale=0.283] (958) at (3.16647, -5) {};
		\node [style=none] (960) at (5.66647, -3.5) {};
		\node [style=none] (961) at (5.66647, -3.5) {};
		\node [style=element] (963) at (0.91647, -3.5) {$The~students$};
		\node [style=element] (964) at (3.16647, -3.5) {$read$};
		\node [style=element] (965) at (5.41647, -3.5) {$the~books$};
		\node [style=none] (967) at (8.3498, -3.5) {};
		\node [style=none] (968) at (8.3498, -4.5) {};
		\node [style=none] (970) at (12.0998, -3.5) {};
		\node [style=none] (971) at (12.0998, -4.5) {};
		\node [style=none] (976) at (10.0998, -5) {};
		\node [style=element] (979) at (8.3498, -3.5) {$The students$};
		\node [style=element] (980) at (10.0998, -3.5) {$were$};
		\node [style=element] (981) at (12.0998, -3.5) {$learning$};
		\node [style=none] (983) at (5.66647, -3.5) {};
		\node [style=none] (984) at (5.66647, -4.5) {};
		\node [style=none] (985) at (8.1665, -3.5) {};
		\node [style=none] (987) at (3.16647, -5.5) {};
		\node [style=none] (989) at (3.16647, -5.5) {};
		\node [style=none] (990) at (6.66647, -6) {};
		\node [style=none] (991) at (10.1998, -3.5) {};
		\node [style=none] (992) at (10.1998, -5) {};
		\node [style=none] (993) at (10.1998, -5.25) {};
		\node [style=none] (994) at (10.1998, -5.5) {};
		\node [circle, fill=black, scale=0.283] (995) at (10.1998, -5) {};
		\node [style=none] (996) at (10.1998, -5.5) {};
		\node [style=none] (997) at (10.1998, -5.5) {};
		\node [style=none] (998) at (6.69974, -6) {};
		\node [circle, fill=black, scale=0.283] (999) at (6.69974, -6) {};
		\node [style=none] (1000) at (6.68294, -7) {};
		\node [style=process] (1071) at (14.6665, -1.75) {$R_x({\theta_s}_1)$};
		\node [style=process] (1072) at (14.6665, -2.5) {$R_x({\theta_s}_2)$};
		\node [style=process] (1073) at (14.6665, -3.25) {$R_x({\theta_s}_3)$};
		\node [style=state] (1075) at (14.6665, -1) {$0$};
		\node [style=process] (1078) at (16.9165, -1.75) {$R_x({\theta_r}_1)$};
		\node [style=process] (1079) at (16.9165, -2.5) {$R_x({\theta_r}_2)$};
		\node [style=process] (1080) at (16.9165, -3.25) {$R_x({\theta_r}_3)$};
		\node [style=state] (1081) at (16.9165, -1) {$0$};
		\node [style=effect] (1082) at (16.9165, -5.35) {0};
		\node [style=plus] (1083) at (16.9165, -4.6) {};
		\node [style=control] (1085) at (14.6665, -4.6) {};
		\node [style=control] (1092) at (14.6665, -6.35) {};
		\node [style=process] (1111) at (23.9165, -3.75) {$R_x({\theta_w}_1)$};
		\node [style=process] (1112) at (23.9165, -4.5) {$R_x({\theta_w}_2)$};
		\node [style=process] (1113) at (23.9165, -5.25) {$R_x({\theta_w}_3)$};
		\node [style=state] (1114) at (23.9165, -3) {$0$};
		\node [style=process] (1115) at (26.1665, -3.75) {$R_x({\theta_l}_1)$};
		\node [style=process] (1116) at (26.1665, -4.5) {$R_x({\theta_l}_2)$};
		\node [style=process] (1117) at (26.1665, -5.25) {$R_x({\theta_l}_3)$};
		\node [style=state] (1118) at (26.1665, -3) {$0$};
		\node [style=effect] (1119) at (26.1665, -8.6) {0};
		\node [style=plus] (1120) at (26.1665, -7.85) {};
		\node [style=effect] (1122) at (23.9165, -6.85) {0};
		\node [style=effect] (1124) at (21.6665, -9.35) {0};
		\node [style=control] (1125) at (21.6665, -7.85) {};
		\node [style=plus] (1127) at (21.6665, -8.6) {};
		\node [style=control] (1128) at (14.6665, -8.6) {};
		\node [style=plus] (1129) at (23.9165, -6.1) {};
		\node [style=none] (1131) at (14.6665, -9.85) {};
		\node [style=process] (1132) at (19.1665, -1.75) {$R_x({\theta_b}_1)$};
		\node [style=process] (1133) at (19.1665, -2.5) {$R_x({\theta_b}_2)$};
		\node [style=process] (1134) at (19.1665, -3.25) {$R_x({\theta_b}_3)$};
		\node [style=state] (1135) at (19.1665, -1) {$0$};
		\node [style=effect] (1136) at (19.1665, -7.1) {0};
		\node [style=plus] (1137) at (19.1665, -6.35) {};
		\node [style=process] (1138) at (21.6665, -3.75) {$R_x({\theta_s}_1)$};
		\node [style=process] (1139) at (21.6665, -4.5) {$R_x({\theta_s}_2)$};
		\node [style=process] (1140) at (21.6665, -5.25) {$R_x({\theta_s}_3)$};
		\node [style=state] (1141) at (21.6665, -3) {$0$};
		\node [style=control] (1143) at (21.6665, -6.1) {};
	\end{pgfonlayer}
	\begin{pgfonlayer}{edgelayer}
		\draw [in=90, out=-90] (944.center) to (945.center);
		\draw [in=90, out=-90] (947.center) to (948.center);
		\draw [in=90, out=-90] (950.center) to (951.center);
		\draw [in=90, out=-90] (955.center) to (956.center);
		\draw [in=180, out=-90, looseness=0.51] (945.center) to (955.center);
		\draw [in=90, out=-90] (948.center) to (955.center);
		\draw [in=0, out=-90, looseness=0.51] (951.center) to (955.center);
		\draw [in=90, out=-90] (956.center) to (957.center);
		\draw [in=90, out=-90] (967.center) to (968.center);
		\draw [in=90, out=-90] (970.center) to (971.center);
		\draw [in=180, out=-90, looseness=0.51] (968.center) to (976.center);
		\draw [in=0, out=-90, looseness=0.51] (971.center) to (976.center);
		\draw [in=180, out=-90, looseness=0.26] (989.center) to (990.center);
		\draw [in=90, out=-90] (992.center) to (993.center);
		\draw [in=90, out=-90] (991.center) to (992.center);
		\draw [in=90, out=-90] (993.center) to (994.center);
		\draw [in=0, out=-90, looseness=0.26] (997.center) to (998.center);
		\draw (999) to (1000.center);
		\draw (1075) to (1073);
		\draw (1081) to (1080);
		\draw (1083) to (1082);
		\draw (1085) to (1083);
		\draw (1083) to (1080);
		\draw (1073) to (1085);
		\draw (1114) to (1113);
		\draw (1118) to (1117);
		\draw (1120) to (1119);
		\draw (1120) to (1117);
		\draw (1125) to (1120);
		\draw (1128) to (1127);
		\draw (1125) to (1127);
		\draw (1113) to (1129);
		\draw (1129) to (1122);
		\draw (1127) to (1124);
		\draw (1085) to (1092);
		\draw (1092) to (1128);
		\draw (1128) to (1131.center);
		\draw (1135) to (1134);
		\draw (1134) to (1137);
		\draw (1137) to (1136);
		\draw (1092) to (1137);
		\draw (1141) to (1140);
		\draw (1140) to (1125);
		\draw (1143) to (1129);
	\end{pgfonlayer}
\end{tikzpicture}

%% file: emnlp2023.bbl
\begin{thebibliography}{57}
\expandafter\ifx\csname natexlab\endcsname\relax\def\natexlab#1{#1}\fi

\bibitem[{Abbaszade et~al.(2023)Abbaszade, Zomorodi, Salari, and
  Kurian}]{abbaszade2023quantum}
Mina Abbaszade, Mariam Zomorodi, Vahid Salari, and Philip Kurian. 2023.
\newblock \href {http://arxiv.org/abs/2307.16576} {Toward quantum machine
  translation of syntactically distinct languages}.

\bibitem[{Abramsky and Coecke(2008)}]{abramsky2008categorical}
Samson Abramsky and Bob Coecke. 2008.
\newblock \href {http://arxiv.org/abs/0808.1023} {Categorical quantum
  mechanics}.

\bibitem[{Araujo and da~Silva(2020)}]{qensemble1}
Ismael C.~S. Araujo and Adenilton~J. da~Silva. 2020.
\newblock \href {https://doi.org/10.1109/IJCNN48605.2020.9207488} {Quantum
  ensemble of trained classifiers}.
\newblock In \emph{2020 International Joint Conference on Neural Networks
  (IJCNN)}, pages 1--8.

\bibitem[{Batra et~al.(2020)Batra, Zorn, Foil, Minerali, Gawriljuk, Lane, and
  ekins}]{qdrug}
Kushal Batra, Kimberley~M. Zorn, Daniel~H. Foil, Eni Minerali, Victor~O.
  Gawriljuk, Thomas~R. Lane, and sean ekins. 2020.
\newblock \href {https://doi.org/10.26434/chemrxiv.12781232.v1} {Quantum
  machine learning for drug discovery}.
\newblock \emph{ChemRxiv}.

\bibitem[{Brown et~al.(2020)Brown, Mann, Ryder, Subbiah, Kaplan, Dhariwal,
  Neelakantan, Shyam, Sastry, Askell, Agarwal, Herbert-Voss, Krueger, Henighan,
  Child, Ramesh, Ziegler, Wu, Winter, Hesse, Chen, Sigler, Litwin, Gray, Chess,
  Clark, Berner, McCandlish, Radford, Sutskever, and
  Amodei}]{brown2020language}
Tom~B. Brown, Benjamin Mann, Nick Ryder, Melanie Subbiah, Jared Kaplan,
  Prafulla Dhariwal, Arvind Neelakantan, Pranav Shyam, Girish Sastry, Amanda
  Askell, Sandhini Agarwal, Ariel Herbert-Voss, Gretchen Krueger, Tom Henighan,
  Rewon Child, Aditya Ramesh, Daniel~M. Ziegler, Jeffrey Wu, Clemens Winter,
  Christopher Hesse, Mark Chen, Eric Sigler, Mateusz Litwin, Scott Gray,
  Benjamin Chess, Jack Clark, Christopher Berner, Sam McCandlish, Alec Radford,
  Ilya Sutskever, and Dario Amodei. 2020.
\newblock \href {http://arxiv.org/abs/2005.14165} {Language models are few-shot
  learners}.

\bibitem[{Buhrmester et~al.(2019)Buhrmester, Münch, and
  Arens}]{buhrmester2019analysis}
Vanessa Buhrmester, David Münch, and Michael Arens. 2019.
\newblock \href {http://arxiv.org/abs/1911.12116} {Analysis of explainers of
  black box deep neural networks for computer vision: A survey}.

\bibitem[{Chomsky(1957)}]{Chomsky57a}
Noam Chomsky. 1957.
\newblock \emph{Syntactic Structures}.
\newblock Mouton and Co., The Hague.

\bibitem[{Clark and Manning(2016{\natexlab{a}})}]{clark-manning-2016-deep}
Kevin Clark and Christopher~D. Manning. 2016{\natexlab{a}}.
\newblock \href {https://doi.org/10.18653/v1/D16-1245} {Deep reinforcement
  learning for mention-ranking coreference models}.
\newblock In \emph{Proceedings of the 2016 Conference on Empirical Methods in
  Natural Language Processing}, pages 2256--2262, Austin, Texas. Association
  for Computational Linguistics.

\bibitem[{Clark and Manning(2016{\natexlab{b}})}]{clark-manning-2016-improving}
Kevin Clark and Christopher~D. Manning. 2016{\natexlab{b}}.
\newblock \href {https://doi.org/10.18653/v1/P16-1061} {Improving coreference
  resolution by learning entity-level distributed representations}.
\newblock In \emph{Proceedings of the 54th Annual Meeting of the Association
  for Computational Linguistics (Volume 1: Long Papers)}, pages 643--653,
  Berlin, Germany. Association for Computational Linguistics.

\bibitem[{Clark(2021)}]{clark2021old}
Stephen Clark. 2021.
\newblock \href {http://arxiv.org/abs/2109.10044} {Something old, something
  new: Grammar-based ccg parsing with transformer models}.

\bibitem[{Coecke et~al.(2010)Coecke, Sadrzadeh, and Clark}]{Coeckeetal2010}
B.~Coecke, M.~Sadrzadeh, and S.~Clark. 2010.
\newblock {M}athematical {F}oundations for {D}istributed {C}ompositional
  {M}odel of {M}eaning. {L}ambek {F}estschrift.
\newblock \emph{Linguistic Analysis}, 36:345--384.

\bibitem[{Coecke et~al.(2020)Coecke, de~Felice, Meichanetzidis, and
  Toumi}]{coecke2020foundations}
Bob Coecke, Giovanni de~Felice, Konstantinos Meichanetzidis, and Alexis Toumi.
  2020.
\newblock \href {http://arxiv.org/abs/2012.03755} {Foundations for near-term
  quantum natural language processing}.

\bibitem[{Coecke et~al.(2013)Coecke, Grefenstette, and
  Sadrzadeh}]{Coeckeetal2013}
Bob Coecke, Edward Grefenstette, and Mehrnoosh Sadrzadeh. 2013.
\newblock Lambek vs. lambek: Functorial vector space semantics and string
  diagrams for lambek calculus.
\newblock \emph{Ann. Pure and Applied Logic}, 164:1079 -- 1100.

\bibitem[{Coecke and Kissinger(2017)}]{coecke_kissinger_2017}
Bob Coecke and Aleks Kissinger. 2017.
\newblock \href {https://doi.org/10.1017/9781316219317} {\emph{Picturing
  Quantum Processes: A First Course in Quantum Theory and Diagrammatic
  Reasoning}}.
\newblock Cambridge University Press.

\bibitem[{de~Felice et~al.(2021)de~Felice, Toumi, and Coecke}]{de_Felice_2021}
Giovanni de~Felice, Alexis Toumi, and Bob Coecke. 2021.
\newblock \href {https://doi.org/10.4204/eptcs.333.13} {{DisCoPy}: Monoidal
  categories in python}.
\newblock \emph{Electronic Proceedings in Theoretical Computer Science},
  333:183--197.

\bibitem[{Devlin et~al.(2019)Devlin, Chang, Lee, and
  Toutanova}]{devlin2019bert}
Jacob Devlin, Ming-Wei Chang, Kenton Lee, and Kristina Toutanova. 2019.
\newblock \href {http://arxiv.org/abs/1810.04805} {Bert: Pre-training of deep
  bidirectional transformers for language understanding}.

\bibitem[{Frostig et~al.(2018)Frostig, Johnson, and Leary}]{47008}
Roy Frostig, Matthew Johnson, and Chris Leary. 2018.
\newblock \href {https://mlsys.org/Conferences/doc/2018/146.pdf} {Compiling
  machine learning programs via high-level tracing}.

\bibitem[{Ganguly et~al.(2023)Ganguly, Morapakula, and
  Coronado}]{ganguly2023quantum}
Srinjoy Ganguly, Sai~Nandan Morapakula, and Luis Miguel~Pozo Coronado. 2023.
\newblock \href {http://arxiv.org/abs/2305.19383} {Quantum natural language
  processing based sentiment analysis using lambeq toolkit}.

\bibitem[{Grant et~al.(2019)Grant, Wossnig, Ostaszewski, and
  Benedetti}]{Grant_2019}
Edward Grant, Leonard Wossnig, Mateusz Ostaszewski, and Marcello Benedetti.
  2019.
\newblock \href {https://doi.org/10.22331/q-2019-12-09-214} {An initialization
  strategy for addressing barren plateaus in parametrized quantum circuits}.
\newblock \emph{Quantum}, 3:214.

\bibitem[{Grossi et~al.(2022)Grossi, Ibrahim, Radescu, Loredo, Voigt, von
  Altrock, and Rudnik}]{qfraud}
M.~Grossi, N.~Ibrahim, V.~Radescu, R.~Loredo, K.~Voigt, C.~von Altrock, and
  A.~Rudnik. 2022.
\newblock \href {https://doi.org/10.1109/TQE.2022.3213474} {Mixed
  quantum–classical method for fraud detection with quantum feature
  selection}.
\newblock \emph{IEEE Transactions on Quantum Engineering}, 3(01):1--12.

\bibitem[{Havl{\'{\i}}{\v{c}}ek et~al.(2019)Havl{\'{\i}}{\v{c}}ek,
  C{\'{o}}rcoles, Temme, Harrow, Kandala, Chow, and Gambetta}]{Havl_ek_2019}
Vojt{\v{e}}ch Havl{\'{\i}}{\v{c}}ek, Antonio~D. C{\'{o}}rcoles, Kristan Temme,
  Aram~W. Harrow, Abhinav Kandala, Jerry~M. Chow, and Jay~M. Gambetta. 2019.
\newblock \href {https://doi.org/10.1038/s41586-019-0980-2} {Supervised
  learning with quantum-enhanced feature spaces}.
\newblock \emph{Nature}, 567(7747):209--212.

\bibitem[{Holmes et~al.(2022)Holmes, Sharma, Cerezo, and Coles}]{PRXQuantum}
Zo\"e Holmes, Kunal Sharma, M.~Cerezo, and Patrick~J. Coles. 2022.
\newblock \href {https://doi.org/10.1103/PRXQuantum.3.010313} {Connecting
  ansatz expressibility to gradient magnitudes and barren plateaus}.
\newblock \emph{PRX Quantum}, 3:010313.

\bibitem[{Kanovich et~al.(2020)Kanovich, Kuznetsov, Nigam, and
  Scedrov}]{Kanovich2020}
M.~Kanovich, S.~Kuznetsov, V.~Nigam, and A.~Scedrov. 2020.
\newblock \href {https://doi.org/10.1007/978-3-030-51074-9_29} {{Soft
  Subexponentials and Multiplexing}}.
\newblock In \emph{Lecture Notes in Computer Science (including subseries
  Lecture Notes in Artificial Intelligence and Lecture Notes in
  Bioinformatics)}.

\bibitem[{Kaplan et~al.(2020)Kaplan, McCandlish, Henighan, Brown, Chess, Child,
  Gray, Radford, Wu, and Amodei}]{kaplan2020scaling}
Jared Kaplan, Sam McCandlish, Tom Henighan, Tom~B. Brown, Benjamin Chess, Rewon
  Child, Scott Gray, Alec Radford, Jeffrey Wu, and Dario Amodei. 2020.
\newblock \href {http://arxiv.org/abs/2001.08361} {Scaling laws for neural
  language models}.

\bibitem[{Karamlou et~al.(2022)Karamlou, Pfaffhauser, and
  Wootton}]{karamlou2022quantum}
Amin Karamlou, Marcel Pfaffhauser, and James Wootton. 2022.
\newblock \href {http://arxiv.org/abs/2211.00727} {Quantum natural language
  generation on near-term devices}.

\bibitem[{Kartsaklis and Sadrzadeh(2013)}]{kartsaklis-sadrzadeh-2013-prior}
D.~Kartsaklis and M.~Sadrzadeh. 2013.
\newblock Prior disambiguation of word tensors for constructing sentence
  vectors.
\newblock In \emph{Proceedings of the 2013 Conference on Empirical Methods in
  Natural Language Processing}, pages 1590--1601.

\bibitem[{Kartsaklis et~al.(2021{\natexlab{a}})Kartsaklis, Fan, Yeung, Pearson,
  Lorenz, Toumi, de~Felice, Meichanetzidis, Clark, and Coecke}]{lambeq_paper}
Dimitri Kartsaklis, Ian Fan, Richie Yeung, Anna Pearson, Robin Lorenz, Alexis
  Toumi, Giovanni de~Felice, Konstantinos Meichanetzidis, Stephen Clark, and
  Bob Coecke. 2021{\natexlab{a}}.
\newblock \href {https://doi.org/10.48550/ARXIV.2110.04236} {lambeq: An
  efficient high-level python library for quantum nlp}.

\bibitem[{Kartsaklis et~al.(2021{\natexlab{b}})Kartsaklis, Fan, Yeung, Pearson,
  Lorenz, Toumi, de~Felice, Meichanetzidis, Clark, and
  Coecke}]{kartsaklis2021lambeq}
Dimitri Kartsaklis, Ian Fan, Richie Yeung, Anna Pearson, Robin Lorenz, Alexis
  Toumi, Giovanni de~Felice, Konstantinos Meichanetzidis, Stephen Clark, and
  Bob Coecke. 2021{\natexlab{b}}.
\newblock lambeq: {A}n {E}fficient {H}igh-{L}evel {P}ython {L}ibrary for
  {Q}uantum {NLP}.
\newblock \emph{arXiv preprint arXiv:2110.04236}.

\bibitem[{Kerenidis and Luongo(2020)}]{qMNIST}
Iordanis Kerenidis and Alessandro Luongo. 2020.
\newblock \href {https://doi.org/10.1103/PhysRevA.101.062327} {Classification
  of the mnist data set with quantum slow feature analysis}.
\newblock \emph{Phys. Rev. A}, 101:062327.

\bibitem[{Lambek(1988)}]{Lambek1988}
J.~Lambek. 1988.
\newblock \href {https://doi.org/10.1007/978-94-015-6878-4_11}
  {\emph{Categorial and Categorical Grammars}}, pages 297--317. Springer
  Netherlands, Dordrecht.

\bibitem[{Lambek(1958)}]{doi:10.1080/00029890.1958.11989160}
Joachim Lambek. 1958.
\newblock \href {https://doi.org/10.1080/00029890.1958.11989160} {The
  mathematics of sentence structure}.
\newblock \emph{The American Mathematical Monthly}, 65(3):154--170.

\bibitem[{Lee et~al.(2018)Lee, He, and Zettlemoyer}]{lee2018}
Kenton Lee, Luheng He, and Luke Zettlemoyer. 2018.
\newblock \href {https://doi.org/10.18653/v1/N18-2108} {Higher-order
  coreference resolution with coarse-to-fine inference}.
\newblock In \emph{Proceedings of the 2018 Conference of the North {A}merican
  Chapter of the Association for Computational Linguistics: Human Language
  Technologies, Volume 2 (Short Papers)}, pages 687--692, New Orleans,
  Louisiana. Association for Computational Linguistics.

\bibitem[{Levesque et~al.(2012)Levesque, Davis, and Morgenstern}]{levesque2012}
Hector~J. Levesque, Ernest Davis, and Leora Morgenstern. 2012.
\newblock {The winograd schema challenge}.
\newblock In \emph{Proceedings of the International Workshop on Temporal
  Representation and Reasoning}.

\bibitem[{Lewis(2020)}]{lewis2020logical}
Martha Lewis. 2020.
\newblock \href {http://arxiv.org/abs/2005.04929} {Towards logical negation for
  compositional distributional semantics}.

\bibitem[{Lorenz et~al.(2021)Lorenz, Pearson, Meichanetzidis, Kartsaklis, and
  Coecke}]{QnlpInPractice}
Robin Lorenz, Anna Pearson, Konstantinos Meichanetzidis, Dimitri Kartsaklis,
  and Bob Coecke. 2021.
\newblock \href {https://doi.org/10.48550/ARXIV.2102.12846} {Qnlp in practice:
  Running compositional models of meaning on a quantum computer}.

\bibitem[{Macaluso et~al.(2020)Macaluso, Clissa, Lodi, and
  Sartori}]{qensemble2}
Antonio Macaluso, Luca Clissa, Stefano Lodi, and Claudio Sartori. 2020.
\newblock \href {http://arxiv.org/abs/2007.01028} {{Quantum Ensemble for
  Classification}}.

\bibitem[{Manning et~al.(2014)Manning, Surdeanu, Bauer, Finkel, Bethard, and
  McClosky}]{manning-etal-2014-stanford}
Christopher Manning, Mihai Surdeanu, John Bauer, Jenny Finkel, Steven Bethard,
  and David McClosky. 2014.
\newblock \href {https://doi.org/10.3115/v1/P14-5010} {The {S}tanford
  {C}ore{NLP} natural language processing toolkit}.
\newblock In \emph{Proceedings of 52nd Annual Meeting of the Association for
  Computational Linguistics: System Demonstrations}, pages 55--60, Baltimore,
  Maryland. Association for Computational Linguistics.

\bibitem[{McClean et~al.(2018)McClean, Boixo, Smelyanskiy, Babbush, and
  Neven}]{McClean_2018}
Jarrod~R. McClean, Sergio Boixo, Vadim~N. Smelyanskiy, Ryan Babbush, and
  Hartmut Neven. 2018.
\newblock \href {https://doi.org/10.1038/s41467-018-07090-4} {Barren plateaus
  in quantum neural network training landscapes}.
\newblock \emph{Nature Communications}, 9(1).

\bibitem[{McPheat et~al.(2020)McPheat, Wazni, and Sadrzadeh}]{mcpheat2021LACL}
L.~McPheat, H.~Wazni, and M.~Sadrzadeh. 2020.
\newblock Vector space semantics for lambek calculus with soft subexponentials.
\newblock In \emph{Proceedings of the tenth international conference on Logical
  Aspect of Computational Linguistics}.

\bibitem[{Meichanetzidis et~al.(2023)Meichanetzidis, Toumi, de~Felice, and
  Coecke}]{Meichanetzidis_2023}
Konstantinos Meichanetzidis, Alexis Toumi, Giovanni de~Felice, and Bob Coecke.
  2023.
\newblock \href {https://doi.org/10.1007/s42484-023-00097-1} {Grammar-aware
  sentence classification on quantum computers}.
\newblock \emph{Quantum Machine Intelligence}, 5(1).

\bibitem[{Miranda et~al.(2021)Miranda, Yeung, Pearson, Meichanetzidis, and
  Coecke}]{miranda2021quantum}
Eduardo~Reck Miranda, Richie Yeung, Anna Pearson, Konstantinos Meichanetzidis,
  and Bob Coecke. 2021.
\newblock \href {http://arxiv.org/abs/2111.06741} {A quantum natural language
  processing approach to musical intelligence}.

\bibitem[{Nielsen and Chuang(2010)}]{nielsen_chuang_2010}
Michael~A. Nielsen and Isaac~L. Chuang. 2010.
\newblock \href {https://doi.org/10.1017/CBO9780511976667} {\emph{Quantum
  Computation and Quantum Information: 10th Anniversary Edition}}.
\newblock Cambridge University Press.

\bibitem[{Piedeleu and Zanasi(2023)}]{piedeleu2023introduction}
Robin Piedeleu and Fabio Zanasi. 2023.
\newblock \href {http://arxiv.org/abs/2305.08768} {An introduction to string
  diagrams for computer scientists}.

\bibitem[{Preskill(2018)}]{Preskill_2018}
John Preskill. 2018.
\newblock \href {https://doi.org/10.22331/q-2018-08-06-79} {Quantum computing
  in the {NISQ} era and beyond}.
\newblock \emph{Quantum}, 2:79.

\bibitem[{Rahman and Ng(2012)}]{rahman-ng-2012-resolving}
Altaf Rahman and Vincent Ng. 2012.
\newblock \href {https://aclanthology.org/D12-1071} {Resolving complex cases of
  definite pronouns: The winograd schema challenge}.
\newblock In \emph{Proceedings of the 2012 Joint Conference on Empirical
  Methods in Natural Language Processing and Computational Natural Language
  Learning}, pages 777--789, Jeju Island, Korea. Association for Computational
  Linguistics.

\bibitem[{Reimers and Gurevych(2019)}]{reimers2019sentencebert}
Nils Reimers and Iryna Gurevych. 2019.
\newblock \href {http://arxiv.org/abs/1908.10084} {Sentence-bert: Sentence
  embeddings using siamese bert-networks}.

\bibitem[{Ruskanda et~al.(2022)Ruskanda, Rifat~Abiwardani, Al~Bari,
  Arya~Bagaspati, Mulyawan, Syafalni, and Larasati}]{9951286}
Fariska~Z. Ruskanda, Muhammad Rifat~Abiwardani, Muhammad~Akram Al~Bari,
  Kinantan Arya~Bagaspati, Rahmat Mulyawan, Infall Syafalni, and
  Harashta~Tatimma Larasati. 2022.
\newblock \href {https://doi.org/10.1109/QCE53715.2022.00025} {Quantum
  representation for sentiment classification}.
\newblock In \emph{2022 IEEE International Conference on Quantum Computing and
  Engineering (QCE)}, pages 67--78.

\bibitem[{Sadrzadeh et~al.(2013)Sadrzadeh, Clark, and Coecke}]{Sadrzadeh_2013}
M.~Sadrzadeh, S.~Clark, and B.~Coecke. 2013.
\newblock \href {https://doi.org/10.1093/logcom/ext044} {The frobenius anatomy
  of word meanings i: subject and object relative pronouns}.
\newblock \emph{Journal of Logic and Computation}, 23(6):1293--1317.

\bibitem[{Sadrzadeh et~al.(2014)Sadrzadeh, Clark, and Coecke}]{Sadrzadeh_2014}
Mehrnoosh Sadrzadeh, Stephen Clark, and Bob Coecke. 2014.
\newblock \href {https://doi.org/10.1093/logcom/exu027} {The frobenius anatomy
  of word meanings {II}: possessive relative pronouns}.
\newblock \emph{Journal of Logic and Computation}, 26(2):785--815.

\bibitem[{Shepherd and Bremner(2009)}]{Shepherd_2009}
Dan Shepherd and Michael~J. Bremner. 2009.
\newblock \href {https://doi.org/10.1098/rspa.2008.0443} {Temporally
  unstructured quantum computation}.
\newblock \emph{Proceedings of the Royal Society A: Mathematical, Physical and
  Engineering Sciences}, (2105):1413--1439.

\bibitem[{Spall(1998)}]{705889}
J.C. Spall. 1998.
\newblock \href {https://doi.org/10.1109/7.705889} {Implementation of the
  simultaneous perturbation algorithm for stochastic optimization}.
\newblock \emph{IEEE Transactions on Aerospace and Electronic Systems},
  34(3):817--823.

\bibitem[{Steedman(2001)}]{10.7551}
Mark Steedman. 2001.
\newblock \href {https://doi.org/10.7551/mitpress/6591.001.0001} {\emph{{The
  Syntactic Process}}}.
\newblock The MIT Press.

\bibitem[{Stein et~al.(2023)Stein, Christ, Kraus, Mansky, Müller, and
  Linnhoff-Popien}]{stein2023applying}
Jonas Stein, Ivo Christ, Nicolas Kraus, Maximilian~Balthasar Mansky, Robert
  Müller, and Claudia Linnhoff-Popien. 2023.
\newblock \href {http://arxiv.org/abs/2307.11788} {Applying qnlp to sentiment
  analysis in finance}.

\bibitem[{Sutor(2019)}]{sutor2019dancing}
R.S. Sutor. 2019.
\newblock \href {https://books.google.co.uk/books?id=uwAeywEACAAJ}
  {\emph{Dancing with Qubits: How Quantum Computing Works and how it Can Change
  the World}}.
\newblock Expert Insight. Packt Publishing.

\bibitem[{Touvron et~al.(2023)Touvron, Lavril, Izacard, Martinet, Lachaux,
  Lacroix, Rozière, Goyal, Hambro, Azhar, Rodriguez, Joulin, Grave, and
  Lample}]{touvron2023llama}
Hugo Touvron, Thibaut Lavril, Gautier Izacard, Xavier Martinet, Marie-Anne
  Lachaux, Timothée Lacroix, Baptiste Rozière, Naman Goyal, Eric Hambro,
  Faisal Azhar, Aurelien Rodriguez, Armand Joulin, Edouard Grave, and Guillaume
  Lample. 2023.
\newblock \href {http://arxiv.org/abs/2302.13971} {Llama: Open and efficient
  foundation language models}.

\bibitem[{Wazni et~al.(2022)Wazni, Lo, McPheat, and
  Sadrzadeh}]{wazni2022quantum}
Hadi Wazni, Kin~Ian Lo, Lachlan McPheat, and Mehrnoosh Sadrzadeh. 2022.
\newblock \href {http://arxiv.org/abs/2208.05393} {A quantum natural language
  processing approach to pronoun resolution}.

\bibitem[{Yeung and Kartsaklis(2021)}]{yeung2021ccgbased}
Richie Yeung and Dimitri Kartsaklis. 2021.
\newblock \href {http://arxiv.org/abs/2105.07720} {A ccg-based version of the
  discocat framework}.

\end{thebibliography}
